\documentclass[letterpaper, 10 pt, conference]{ieeeconf}  

\IEEEoverridecommandlockouts                              
\usepackage{amsmath} 
\usepackage{amssymb}  
\usepackage{subcaption}

\usepackage{mathtools}
\newtheorem{theorem}{\bfseries Theorem}[section]
\newtheorem{lemma}{\bfseries Lemma}[section]
\newtheorem{assumption}{\bfseries Assumption}[section]
\newtheorem{definition}{\bfseries Definition}[section]
\newtheorem{proposition}{\bfseries Proposition}[section]
\usepackage{algorithm}
\usepackage{algpseudocode}
\usepackage{xcolor}
\usepackage{hyperref}
\hypersetup{breaklinks=true,colorlinks=true,linkcolor=black,citecolor=black}
\usepackage[noabbrev,capitalise,nameinlink]{cleveref}
\usepackage[sort]{cite}
\crefname{assumption}{Assumption}{Assumptions}
\allowdisplaybreaks

\title{\LARGE \bf
Finite-Time Error Analysis of Soft Q-Learning: Switching System Approach
}

\author{Narim Jeong and Donghwan Lee
\thanks{N. Jeong and D. Lee are with the Department of Electrical Engineering, Korea Advanced Institute of Science and Technology (KAIST), Daejeon, 34141, South Korea
        {\tt\small \{nrjeong, donghwan\}@kaist.ac.kr}}
}

\begin{document}

\maketitle
\thispagestyle{empty}
\pagestyle{empty}

\begin{abstract}
Soft Q-learning is a variation of Q-learning designed to solve entropy regularized Markov decision problems where an agent aims to maximize the entropy regularized value function. Despite its empirical success, there have been limited theoretical studies of soft Q-learning to date. This paper aims to offer a novel and unified finite-time, control-theoretic analysis of soft Q-learning algorithms. We focus on two types of soft Q-learning algorithms: one utilizing the log-sum-exp operator and the other employing the Boltzmann operator. By using dynamical switching system models, we derive novel finite-time error bounds for both soft Q-learning algorithms. We hope that our analysis will deepen the current understanding of soft Q-learning by establishing connections with switching system models and may even pave the way for new frameworks in the finite-time analysis of other reinforcement learning algorithms.
\end{abstract}

\section{INTRODUCTION}
In recent years, a range of reinforcement learning (RL) algorithms have been introduced, effectively addressing numerous complex challenges~\cite{van2016deep, schulman2015trust, schulman2017proximal}. In particular, Q-learning~\cite{watkins1992q} stands as a fundamental RL algorithm, with its convergence extensively examined through decades of research~\cite{jaakkola1994convergence, borkar2000ode, even2003learning, beck2012error, qu2020finite, wainwright2019stochastic, lee2023discrete}. Soft Q-learning~\cite{haarnoja2017reinforcement} is a modified Q-learning for solving entropy regularized Markov decision problems, where an agent aims to maximize the entropy regularized value function. Due to its favorable properties, such as smoothness and effectiveness in exploration, soft Q-learning has demonstrated its efficacy across a range of applications, which triggered a surge of interest in this direction~\cite{neu2017unified, nachum2017bridging, schulman2017equivalence,asadi2017alternative, haarnoja2018soft, pan2019reinforcement,geist2019theory,liang2021temporal,cai2023neural,smirnova2019convergence,ying2022dual}. Despite its empirical success, to the best of the authors' knowledge, theoretical research on the convergence of soft Q-learning has been significantly limited to date.

Motivated by the above discussions, we present a novel finite-time error analysis for two variants of soft Q-learning algorithms: one based on the log-sum-exp (LSE) operator and the other on the Boltzmann operator. Inspired by the switching system technique applied in standard Q-learning analysis~\cite{lee2023discrete}, we initially model the update rule of soft Q-learning as discrete-time nonlinear dynamic system models. To address the complex nonlinearity of the systems, we develop the upper and lower bounds for both LSE and Boltzmann operators. This approach allows us to devise two comparison systems: a \textit{lower comparison system} and an \textit{upper comparison system}, which play key roles in our analysis. The former underestimates, while the latter overestimates, the trajectories of the original soft Q-learning process. Then, a finite-time error bound for each comparison system is derived and used to prove the finite-time convergence of soft Q-learning dynamics.

\vspace{0.3cm}
\paragraph*{Related works}
Soft Q-learning has been introduced in~\cite{haarnoja2017reinforcement} in order to solve the entropy regularized Markov decision problem. It has been extended to soft actor-critic methods~\cite{haarnoja2018soft}, which have shown considerable empirical success. Afterwards, there have been significant advances in a number of works. For instance,~\cite{neu2017unified} proposed a unified framework to understand and analyze various entropy regularized methods in terms of the mirror decent methods,~\cite{nachum2017bridging} established a new link between the value and policy-based RLs in terms of the entropy regularized problems, and~\cite{schulman2017equivalence} proved an equivalence between Q-learning and policy gradient approaches in entropy-regularized RLs. The convergence of a class of entropy regularized dynamic programming algorithms was investigated in~\cite{smirnova2019convergence}, alternative softmax operators were studied in~\cite{asadi2017alternative}, and a version of soft Q-learning was developed and analyzed in~\cite{liang2021temporal} to control the overestimation bias in Q-learning. Moreover,~\cite{geist2019theory} suggested a general theory of regularized Markov decision problems based on the regularized Bellman operator and Legendre-Fenchel transform, which enabled error propagation analysis of various algorithms as special cases, and~\cite{ying2022dual} addressed entropy-regularized constrained Markov decision problems with Lagrangian duality and the accelerated dual descent method.

Despite the significant advances in theoretical analysis of soft Q-learning and entropy regularized problems, convergence analysis of soft Q-learning has received comparatively little attention. In particular, convergence analysis of soft Q-learning or its variants has been studied in~\cite{cai2023neural,pan2019reinforcement}. In~\cite{cai2023neural}, a neural soft Q-learning with the LSE operator has been proposed with its global convergence analysis. Furthermore,~\cite{pan2019reinforcement} suggested a soft Q-learning based on the Boltzmann operator with tabular settings and proved its convergence.

\vspace{0.3cm}
\paragraph*{Contribution}
Our approach aims to complement the previous works by proposing a novel strategy: switching system models of soft Q-learning with control theoretic concepts. We provide finite-time (non-asymptotic) convergence for soft Q-learning with both LSE and the Boltzmann operators using constant step size, which cover different scenarios compared to~\cite{cai2023neural,pan2019reinforcement}. In particular, although~\cite{cai2023neural} considered more general function approximation scenarios, it adopted stricter assumptions on the step-size, exploration, and error terms in order to deal with the global convergence with nonlinearity in their function approximation. Moreover, their algorithm includes neural network function approximations with additional projection and averaging steps. Therefore, it is still meaningful to examine the finite-time analysis of naive soft Q-learning algorithms, which is lacking in the literature, with milder conditions to complement the existing analysis and improve our understanding. Furthermore,~\cite{pan2019reinforcement} only provides asymptotic convergence with diminishing step sizes. Our analysis provides a unified framework to cover both the Boltzmann and LSE operators, while~\cite{pan2019reinforcement} only considers the Boltzmann operator.

\section{PRELIMINARIES}

\subsection{Markov Decision Problem}
We consider a Markov decision process (MDP) that has the state-space $\mathcal{S} := \{1, 2, \ldots, |\mathcal{S}|\}$ and action-space $\mathcal{A} := \{1, 2, \ldots, |\mathcal{A}|\}$, where $|\mathcal{S}|$ and $|\mathcal{A}|$ indicate the cardinality of $\mathcal{S}$ and $\mathcal{A}$, respectively. Based on a current state $s_k \in \mathcal{S}$ in time-step $k$, an agent selects an action $a_k \in \mathcal{A}$ with a policy $\pi$. Then, with the state transition probability $P(s_k'|s_k,a_k)$, the state $s_k$ transits to the next state $s_k' \in \mathcal{S}$, which incurs a reward $r_k := r(s_k, a_k, s_k')$.

Given an MDP, the Markov decision problem is to discover an optimal policy $\pi^*$ that maximizes the cumulative discounted rewards, i.e.,
\[\pi^* := \mathrm{argmax}_{\pi \in \Theta} \mathbb{E} \left[\sum_{k=0}^{\infty} \gamma^k r_k \Bigg| \pi \right].\]
Here, $\Theta$ is the set of all possible policies, $\mathbb{E}[\cdot|\pi]$ is the expectation with regard to the policy $\pi$, and $\gamma \in [0, 1)$ is the discount factor.

The Q-function represents the expected cumulative rewards when action $a$ is taken at state $s$ and a given policy $\pi$ is followed afterwards, denoted as
\[ Q^\pi(s, a) := \mathbb{E} \left[\sum_{k=0}^{\infty} \gamma^k r_k \Bigg| s_0=s, a_0=a, \pi \right], s \in \mathcal{S}, a \in \mathcal{A}. \]
Moreover, the optimal Q-function refers to the Q-function produced by the optimal policy $\pi^*$, defined as $Q^*(s,a) := Q^{\pi^*}(s,a)$ with $s \in \mathcal{S}, a \in \mathcal{A}$. After the optimal Q-function $Q^*$ is determined, the optimal policy can be obtained as the greedy policy $\pi^*(s) = \arg\max_{a \in \mathcal{A}} Q^*(s,a)$.

\subsection{Switching System}
A general nonlinear system~\cite{khalil2002nonlinear} is defined as $x_{k+1}=f(x_k)$ with $x_0 \in \mathbb{R}^n$, where $x_k \in \mathbb{R}^n$ is the state, $k \in \{0, 1, 2, \ldots\}$ is the time-step, and $f: \mathbb{R}^n \rightarrow \mathbb{R}^n$ is a nonlinear function.

The switching system~\cite{liberzon2003switching} is a particular form of the nonlinear system, which uses switching signals to operate across several subsystems. In this paper, we consider the following particular switching system, known as the \textit{affine switching system}:
\[x_{k+1}=A_{\sigma_k}x_k+b_{\sigma_k}, \; x_0 \in \mathbb{R}^n, \; k \in \{0, 1, \ldots\},\]
where $\sigma_k \in \mathcal{M} := \{1, 2, \ldots, M\}$ is the switching signal, $A_{\sigma_k} \in \mathbb{R}^{n \times n}$ is the subsystem matrix, and $b_{\sigma_k} \in \mathbb{R}^n$ is the additional input vector. The switching signal $\sigma_k$ can be generated arbitrarily or according to policy, and $A_{\sigma_k}$ and $b_{\sigma_k}$ are changed in accordance with $\sigma_k$. Its stabilization is significantly difficult due to the additional input $b_{\sigma_k}$.

\begin{algorithm} [t]
   \caption{Variants of soft Q-learning}
   \label{alg1}
    \begin{algorithmic}
       \State Initialize $Q(s,a)$ for all $s \in \mathcal{S}$ and $a \in \mathcal{A}$
       \State Choose the operator $h(Q(s,\cdot))$ from~\eqref{h_lse} or~\eqref{h_bz}
       \State Initialize parameter $\beta$ of $h(Q(s,\cdot))$
       \State Observe $s_0$ in accordance with the initial state distribution
          \For{$k=0,1,\ldots$}
            \State Take action under the behavior policy: $a_k \sim \pi_b(\cdot|s_k)$
            \State Observe $s_k' \sim P(\cdot|s_k,a_k)$ and  $r_k = r(s_k,a_k,s_k')$
            \State Update $Q$-function:
            \begin{align*}
                \hspace{0.5cm} Q(s_k,a_k) \leftarrow& Q(s_k,a_k) \\
                &+ \alpha \{r_k +\gamma h(Q(s_k',\cdot)) - Q(s_k,a_k)\}
            \end{align*}
            \State Update $s_k \leftarrow s_k'$
          \EndFor
    \end{algorithmic}
\end{algorithm}

\subsection{Soft Q-learning}
Through this paper, we consider the two different versions of soft Q-learning presented in~\cref{alg1}. The only difference between them is the operator $h$ that can handle various cases in the Q-function update equation: For any $v \in {\mathbb R}^{|{\cal A}|}$,~\cref{alg1} uses the LSE operator if $h(v)$ is defined as
\begin{equation} \label{h_lse}
    h_{LSE}^{\beta}(v) : = \frac{1}{\beta }\ln \left( {\sum\limits_{a \in {\cal A}} {\exp } (v(a)\beta )} \right),
\end{equation}
whereas the Boltzmann operator if $h(v)$ is defined as
\begin{equation} \label{h_bz}
    h_{Boltz}^{\beta}(v) := \frac{\sum_{a \in \mathcal{A}} v(a) \exp(v(a)\beta)}{\sum_{u \in \mathcal{A}} \exp(v(u)\beta )}.
\end{equation}
For convenience,~\cref{alg1} with~\eqref{h_lse} will be called \textit{LSE soft Q-learning}, and with~\eqref{h_bz} will be called \textit{Boltzmann soft Q-learning}. Here, the parameter $\beta > 0$ determines the sharpness of the operators, implying that a larger $\beta$ results in a sharper approximation of the max operator. Soft Q-learning utilizes those operators to approximate the max operator in the original Q-learning.

The LSE operator is known to be a non-expansion and so has a unique fixed point~\cite{dai2018sbeed}. However, it is known that the Boltzmann operator is not non-expansive; it can have multiple fixed points, and it is at risk of misbehaving~\cite{asadi2017alternative}. Depending on the properties of each operator, it might be challenging to examine the convergence of the algorithms. For analysis, we examine the algorithms in the tabular domains in order to adjust them to the switching system framework.

\subsection{Assumptions and Definitions}
We assume that the state-action pairs $\{(s_k, a_k)\}_{k=0}^{\infty}$ are selected independently and identically distributed under the behavior policy $\pi_b$, and MDP is ergodic, which is common in the reinforcement learning field. Moreover, we suppose that the behavior policy $\pi_b$ is time-invariant, so that the stationary state-action distribution $d(s,a)=p(s)\pi_b(a|s)$ is fixed with a stationary state distribution $p(s)$. Since the stationary state-action distribution value is constant, we can avoid additional nonlinearity and probabilistic dependency in the switching system.

In addition, we adopt the following assumptions for analytical convenience.
\vspace{0.3cm}
\begin{assumption}\label{as: 1}
Throughout this paper, we consider the following assumptions:
\begin{enumerate}
    \item The step size $\alpha \in (0,1)$ is a constant.
    \item The reward is unit-bounded as follows:
    \[\max_{(s,a,s^{\prime}) \in \mathcal{S} \times \mathcal{A} \times \mathcal{S}} |r(s,a,s^{\prime})| =: R_{max} \le 1.\]
    \item The initial iterate of the Q-functions $Q_0^{LSE}$ and $Q_0^{Boltz}$ are unit-bounded as follows:
    \[\left\|Q_0^{LSE}\right\|_{\infty} \le 1 \;\text{and}\; \left\|Q_0^{Boltz}\right\|_{\infty} \le 1.\]
    \item $d(s,a)>0$ for all $(s,a)\in {\cal S} \times {\cal A}.$
\end{enumerate}
\end{assumption} \vspace{0.3cm}

We note that the assumption $\alpha \in (0,1)$ is essential for our main theoretical analysis, while the unit-bounded rewards and initial parameters are only for analytic convenience without loss of generality.

Moreover, the terms listed below will be used frequently in this work.
\vspace{0.3cm}
\begin{definition} \label{def1}
    The maximum and minimum state-action visit probabilities are defined as
    \[ d_{max} := \max_{(s,a) \in \mathcal{S} \times \mathcal{A}} d(s,a) \in (0,1),\]
    \[ d_{min} := \min_{(s,a) \in \mathcal{S} \times \mathcal{A}} d(s,a) \in (0,1),\]
    where $d(s,a)$ is the stationary state-action distribution.
\end{definition} \vspace{0.3cm}
Here, $d(s,a)=p(s)\pi_b(a|s)$ can be seen as the stationary state-action distribution under the behavior policy $\pi_b$ for practical extensions. However, note that our main analysis is valid for any probability distribution $d(s,a)$ over the state-action pair.
\vspace{0.3cm}
\begin{definition} \label{def3}
    The exponential decay rate is defined as
    \[\rho := 1-\alpha d_{min}(1-\gamma) \in (0,1).\]
\end{definition} \vspace{0.3cm}

Furthermore, we employ dynamic system models for our main finite-time error analysis. Some matrices and vectors listed below will be useful for notational and analytical simplicity.
\vspace{0.3cm}
\begin{definition} \label{def8}
We consider the following vector and matrix notations:
    \[P :=
      \begin{bmatrix} 
        P_1 \\
        \vdots \\
        P_{|\mathcal{A}|} \\
      \end{bmatrix},
    R :=
      \begin{bmatrix} 
        R_1 \\
        \vdots \\
        R_{|\mathcal{A}|} \\
      \end{bmatrix},\]
    \[Q :=
      \begin{bmatrix} 
        Q(\cdot,1) \\
        \vdots \\
        Q(\cdot,|\mathcal{A}|) \\
      \end{bmatrix},
    \Pi^{\pi} := \left[ {\begin{array}{*{20}{c}}
        {\pi {{(1)}^T} \otimes e_1^T}\\
         \vdots \\
        {\pi {{(|{\cal S}|)}^T} \otimes e_{|{\cal S}|}^T}
        \end{array}} \right],\]
    \[D_a :=
      \begin{bmatrix} 
        d(1,a) \\
        & \ddots \\
        & & d(|\mathcal{S}|, a) \\
      \end{bmatrix},
    D :=
      \begin{bmatrix} 
        D_1 \\
        & \ddots \\
        & & D_{|\mathcal{A}|} \\
      \end{bmatrix}\]
    , where $P_a=P(\cdot|a, \cdot) \in \mathbb{R}^{|\mathcal{S}| \times |\mathcal{S}|}$ is the state transition probability matrix when choosing action $a\in \mathcal{A}$, $R_a(s) = \mathbb{E}[r(s,a,s^{\prime})|s,a]$, and $Q(\cdot, a) \in \mathbb{R}^{|\mathcal{S}|}$ is a vector enumerating Q-function values corresponding to action $a \in \mathcal{A}$. In addition, $\pi$ is a given policy, $\pi(s) \in \mathbb{R}^{|\mathcal{A}|}$ is the probability vector of selecting action at the state $s \in {\cal S}$, $\otimes$ is the Kronecker product, and $e_s \in \mathbb{R}^{|\mathcal{S}|}$ is the $s$th basis vector (except for the $s$th component, which is 1, every element is 0). Note that $P \in \mathbb{R}^{|\mathcal{S}||\mathcal{A}| \times |\mathcal{S}|}$, $R \in \mathbb{R}^{|\mathcal{S}||\mathcal{A}|}$, $Q \in \mathbb{R}^{|\mathcal{S}||\mathcal{A}|}$, $\Pi^{\pi} \in \mathbb{R}^{|\mathcal{S}| \times |\mathcal{S}||\mathcal{A}|}$, and $D \in \mathbb{R}^{|\mathcal{S}||\mathcal{A}| \times |\mathcal{S}||\mathcal{A}|}$.
\end{definition} \vspace{0.3cm}

From these, we can express the transition probability matrix of the state-action pair with the policy $\pi$ as $P\Pi^\pi \in \mathbb{R}^{|\mathcal{S}||\mathcal{A}| \times |\mathcal{S}||\mathcal{A}|}$ and the single Q-function value as $Q(s, a) = (e_a \otimes e_s)^TQ$. Here, $e_s \in \mathbb{R}^{|\mathcal{S}|}$ is the $s$th basis vector, and $e_a \in \mathbb{R}^{|\mathcal{A}|}$ is the $a$th basis vector.

Let us define the greedy policy $\pi_{Q,\max}(s) := \mathrm{argmax}_{a \in \mathcal{A}} Q(s,a) \in \mathcal{A}$ and the vector ${\vec \pi }_{Q,\max}(s): = e_{{\pi_{Q,\max}}(s)}$ that converts the deterministic greedy policy $\pi_{Q,\max}$ into the stochastic vector. Then, by using~\cref{def8}, we can show that
\[\Pi^{\vec \pi_{Q,\max}} Q = \left[ {\begin{array}{*{20}{c}}
{{\max }_{a \in A}}Q(1,a)\\
\vdots \\
{{\max }_{a \in A}}Q(|\mathcal{S}|,a)
\end{array}} \right] \in {\mathbb R}^{|{\cal S}|}.\]

Hereafter, we will write 
\[{\Pi ^{{{\vec \pi }_{Q,\max}}}}: = \Pi _Q^{\max}\]
for notational convenience.

\section{SWITCHING SYSTEM FOR Q-LEARNING}
We revisit the switching system framework for Q-learning in~\cite{lee2023discrete} because our main analysis borrows some concepts from this technique. In particular, based on~\crefrange{def1}{def8}, the Q-learning update equation
\begin{equation}\label{eq: Q-learning update equation}
\begin{aligned}
    Q_{k+1}(s_k,a_k) =& Q_k(s_k,a_k) +\alpha \bigg\{r_k \\
    &+ \gamma \max_{u \in \mathcal{A}}Q_k(s_k',u) -Q_k(s_k,a_k) \bigg\}
\end{aligned}
\end{equation}
can be described by
\begin{equation}\label{eq: Q-learning switching}
    Q_{k+1}=Q_k+\alpha \left( DR+\gamma DP\Pi_{Q_k}^{\max}Q_k-DQ_k+w_k \right),
\end{equation}
where
\begin{equation*}
\begin{aligned}
    w_k =& (e_{a_k} \otimes e_{s_k})r_k + \gamma(e_{a_k} \otimes e_{s_k})(e_{s_k'})^T \Pi_{Q_k}^{\max}Q_k \\
    &- (e_{a_k} \otimes e_{s_k})(e_{a_k} \otimes e_{s_k})^T Q_k \\
    &- \left( DR+\gamma DP\Pi_{Q_k}^{\max}Q_k-DQ_k \right).
\end{aligned}
\end{equation*}
Here, $(s_k, a_k, r_k, s_k')$ is selected at the $k$th time-step, each element of $DP$ is a joint distribution of $(s,a,s^{\prime}) \in \mathcal{S} \times \mathcal{A} \times \mathcal{S}$, and $\Pi_{Q_k}^{\max}Q_k$ plays a role of the max operator in~\eqref{eq: Q-learning update equation}.

Using the optimal Bellman equation $(\gamma DP\Pi _{Q^*}^{\max}-D)Q^*+DR=0$,~\eqref{eq: Q-learning switching} can be additionally expressed as
\begin{equation}\label{eq: stochastic affine switching system}
  Q_{k+1}-Q^*=A_{Q_k}(Q_k-Q^*)+b_{Q_k}+\alpha w_k
\end{equation}
with
\[A_{Q_k} := I+\alpha \left(\gamma DP\Pi_{Q_k}^{\max}-D \right),\]
\[b_{Q_k} := \alpha \gamma DP \left(\Pi_{Q_k}^{\max}-\Pi _{Q^*}^{\max} \right)Q^*.\]
Then,~\eqref{eq: stochastic affine switching system} can be seen as a \textit{stochastic affine switching system} with an additional affine term $b_{Q_k}$ and a stochastic noise $w_k$, where $A_{Q_k}$ and $b_{Q_k}$ change in response to the greedy policy.

With this switching system model, determining the convergence of Q-learning can be regarded as evaluating the stability of~\eqref{eq: stochastic affine switching system}. However, additional affine and stochastic noise terms in~\eqref{eq: stochastic affine switching system} make it difficult to establish the stability. To overcome this problem, we can apply two simpler comparison systems called \textit{lower comparison system} and \textit{upper comparison system}, whose trajectories hold the lower and upper bounds of the Q-learning iterate $Q_k$, respectively.

\section{FINITE-TIME ANALYSIS OF LSE SOFT Q-LEARNING}
In this section, we study the finite-time error analysis of LSE soft Q-learning. To this end, a nonlinear dynamical system model of LSE soft Q-learning is developed first. Then, the corresponding lower and upper comparison systems are obtained, which are simpler linear or switching systems. Using these comparison systems, a finite-time error bound of LSE soft Q-learning can be derived.

\subsection{Nonlinear System Representation of LSE Soft Q-learning}
To begin with, LSE soft Q-learning can be written as
\begin{equation*}
\begin{aligned}
    Q_{k + 1}^{LSE} =& Q_k^{LSE} + \alpha \Big\{ DR + \gamma DP H_{LSE}^{\beta}\left(Q_k^{LSE}\right) \\
    &- DQ_k^{LSE} + {w_k^{LSE}} \Big\},
\end{aligned}
\end{equation*}
where 
\begin{align*}
    H_{LSE}^{\beta}(Q_k) := 
    \left[ {\begin{array}{*{20}{c}}
        h_{LSE}^{\beta}(Q_k(1,\cdot)) \\
        \vdots \\
        h_{LSE}^{\beta}(Q_k(|\mathcal{S}|,\cdot))
    \end{array}} \right],
\end{align*}

\begin{equation} \label{eq: lse noise}
\begin{aligned}
    & w_k^{LSE} \\
    =& (e_{a_k} \otimes e_{s_k})r_k + \gamma(e_{a_k} \otimes e_{s_k})(e_{s_k'})^T  H_{LSE}^{\beta}\left(Q_k^{LSE}\right) \\
    &- (e_{a_k} \otimes e_{s_k})(e_{a_k} \otimes e_{s_k})^T Q_k^{LSE} \\
    &- \left\{DR+\gamma DP H_{LSE}^{\beta}\left(Q_k^{LSE}\right) - DQ_k^{LSE}\right\}.
\end{aligned}
\end{equation}

This system is nonlinear, and its stability is relevantly hard to verify. Therefore, instead of directly dealing with this nonlinear system, we will develop the comparison systems in the next subsection, whose structures are much simpler compared to the original system. We will use the proposition listed below, with proof supplied in Appendix.
\vspace{0.3cm}
\begin{proposition} \label{prop3.1}
For any $Q \in {\mathbb R}^{|\mathcal{S}||\mathcal{A}|}$, we have
\[ \max_{a \in \mathcal{A}}Q(s,a) \le h_{LSE}^{\beta}(Q(s,\cdot)) \le \max_{a \in \mathcal{A}}Q(s,a) + \frac{\ln(|\mathcal{A}|)}{\beta} \]
and hence
\[ {\Pi_Q^{\max}}Q \le H_{LSE}^{\beta}(Q) \le {\Pi_Q^{\max}}Q + \frac{{\ln (|\mathcal{A}|)}}{\beta }{\bf{1}}, \]
where ``$\le$'' denotes the element-wise inequality and $\textbf{1}$ is a column vector whose all elements are equal to 1.
\end{proposition} \vspace{0.3cm}

In addition, we employ the following lemmas, which are significant in this study. 
\vspace{0.3cm}
\begin{lemma}[\cite{gosavi2006boundedness}] \label{lem3.6}
    The value of $Q^*$ is bounded as
    \[\|Q^*\|_{\infty} \le \frac{1}{1-\gamma}.\]
\end{lemma} \vspace{0.3cm}

\begin{lemma}[\cite{lee2023discrete}] \label{lem3.2}
    For any $Q \in \mathbb{R}^{|\mathcal{S}||\mathcal{A}|}$,
    \[\|A_Q\|_{\infty} \le \rho,\]
    where $\|A\|_{\infty} := \max_{1 \le i \le m} \sum_{j=1}^n |A_{ij}|$, $A_{ij}$ is the element of $A$ in the $i$th row and $j$th column, and $\rho$ is defined in~\cref{def3}.
\end{lemma} \vspace{0.3cm}

\subsection{Lower Comparison System of LSE Soft Q-learning}
Let us consider the linear system
\begin{equation}\label{eq: soft q-learning lower comparison system}
    Q_{k+1}^L-Q^*=A_{Q^*}\left(Q_k^L-Q^*\right)+\alpha w_k^{LSE}
\end{equation}
with
\[A_{Q^*} := I+\alpha \left(\gamma DP\Pi _{Q^*}^{\max}-D \right)\]
and the stochastic noise~\eqref{eq: lse noise}, which is called \textit{lower comparison system of LSE soft Q-learning}. As the following proposition shows, this system can serve as a lower bound for the LSE soft Q-learning iteration.

\vspace{0.3cm}
\begin{proposition}
    Suppose $Q_0^L - {Q^*} \le Q_0^{LSE} - {Q^*}$, where $Q_0^L-Q^* \in \mathbb{R}^{|\mathcal{S}||\mathcal{A}|}$ and ``$\le$" is the element-wise inequality. Then, for all $k \ge 0$, 
    \[Q_k^L - {Q^*} \le Q_k^{LSE} - {Q^*}.\]
\end{proposition} \vspace{0.3cm}

\begin{proof}
To apply the induction argument, assume that $Q_k^L - {Q^*} \le Q_k^{LSE} - {Q^*}$ holds for some $k \ge 0$. Then, we have
\begin{align*}
    & Q_{k + 1}^{LSE} - {Q^*}\\
    =& Q_k^{LSE} - {Q^*} \\
    &+ \alpha \left\{ DR + \gamma DPH_{LSE}^{\beta}\left(Q_k^{LSE}\right) - DQ_k^{LSE} + w_k^{LSE}\right\} \\
    \ge& Q_k^{LSE} - {Q^*} \\
    &+ \alpha \left(DR + \gamma DP{\Pi _{Q_k^{LSE}}^{\max}}Q_k^{LSE} - DQ_k^{LSE} + w_k^{LSE}\right) \\ 
    \ge& Q_k^{LSE} - {Q^*} \\
    &+ \alpha \left(DR + \gamma DP{\Pi _{Q^*}^{\max}}Q_k^{LSE} - DQ_k^{LSE} + w_k^{LSE}\right) \\
    =& Q_k^{LSE} - {Q^*} - \alpha \left(\gamma DP{\Pi _{Q^*}^{\max}} - D \right)Q^* \\
    &+ \alpha \left(\gamma DP{\Pi _{Q^*}^{\max}}Q_k^{LSE} - DQ_k^{LSE} + w_k^{LSE}\right) \\
    =& \left\{I+ \alpha \left(\gamma DP{\Pi _{Q^*}^{\max}} - D \right)\right\}\left(Q_k^{LSE} - {Q^*}\right) + \alpha w_k^{LSE} \\
    \ge& \left\{I+ \alpha \left(\gamma DP{\Pi _{Q^*}^{\max}} - D \right)\right\}\left(Q_k^L - {Q^*}\right) + \alpha w_k^{LSE} \\
    =& Q_{k+1}^L - {Q^*},
\end{align*}
where~\cref{prop3.1} is used in the first inequality and the optimal Bellman equation is utilized in the second equality. The last inequality relies on the hypothesis $Q_k^L - {Q^*} \le Q_k^{LSE} - {Q^*}$ and the fact that all elements of $I+ \alpha \left(\gamma DP{\Pi _{Q^*}^{\max}} - D\right)$ are non-negative. This completes the proof.
\end{proof} \vspace{0.3cm}

Taking inspiration from \cite{lee2024final}, we can analyze the lower comparison system of LSE soft Q-learning by examining the linear recursion of the autocorrelation matrix of~\eqref{eq: soft q-learning lower comparison system}. Once 
\[X_k := \mathbb{E}\left[\left(Q_{k+1}^L - {Q^*}\right)\left(Q_{k+1}^L - {Q^*}\right)^T\right],\]
\begin{equation} \label{eq: W_k}
    W_k := \mathbb{E}\left[\left(w_k^{LSE}\right)\left(w_k^{LSE}\right)^T\right],
\end{equation}
and $W_k=W_k^T \succeq 0$ are defined, it can be expressed as
\begin{equation} \label{autocorrelation matrix}
    X_{k+1}=A_{Q^*}X_kA_{Q^*}^T+\alpha^2W_k
\end{equation}
using the next lemma:
\vspace{0.3cm}
\begin{lemma} \label{lem w lse expectation}
    For all $k \ge 0$, we have
    \[\mathbb{E}\left[w_k^{LSE}\right]=0.\]
\end{lemma} \vspace{0.3cm}
The proof of~\cref{lem w lse expectation} is available in Appendix.

For investigating~\eqref{autocorrelation matrix}, the following lemma shows the trace of $X_k$, $\text{tr}(X_k)$, which is crucial in the main result.
\vspace{0.3cm}
\begin{lemma} \label{lem3.12}
For all $k \ge 0$, the trace of $X_k$ is bounded as
\begin{align*}
    \text{tr}(X_k) \le&  \frac{6\alpha \left(\ln (|\mathcal{A}|)+\beta\right)^2|\mathcal{S} \times \mathcal{A}|^2}{\beta^2 d_{min}(1-\gamma)^3} \\
     &+ |\mathcal{S} \times \mathcal{A}|^2 \left\|Q_0^L - {Q^*}\right\|_2^2 \rho^{2k}.
\end{align*}
\end{lemma} \vspace{0.3cm}

The proof of~\cref{lem3.12} are available in Appendix.

Then, we can compute a finite-time error bound for the lower comparison system of LSE soft Q-learning.

\vspace{0.3cm}
\begin{theorem}\label{thm3.1}
For any $k \ge 0$, we have
\begin{equation}\label{eq: final iteration lower}
\begin{aligned}
    \mathbb{E}\left[\left\|Q_k^L - Q^*\right\|_2\right] \le& \frac{\sqrt{6} \alpha^\frac{1}{2}\left(\ln (|\mathcal{A}|)+\beta\right) |\mathcal{S} \times \mathcal{A}|}{\beta d_{min}^\frac{1}{2}(1-\gamma)^\frac{3}{2}} \\
    &+ |\mathcal{S} \times \mathcal{A}| \left\|Q_0^L - {Q^*}\right\|_2 \rho^k.
\end{aligned}
\end{equation}
\end{theorem} \vspace{0.3cm}

\begin{proof}
Using the relation
\begin{align*}
    \mathbb{E}\left[\left\|Q_k^L - Q^*\right\|_2^2\right] =& \mathbb{E}\left[\left(Q_k^L - Q^*\right)^T\left(Q_k^L - Q^*\right)\right] \\
    =& \mathbb{E}\left[\text{tr}\left(\left(Q_k^L - Q^*\right)^T\left(Q_k^L - Q^*\right)\right)\right] \\
    =& \mathbb{E}\left[\text{tr}\left(\left(Q_k^L - Q^*\right)\left(Q_k^L - Q^*\right)^T\right)\right] \\
    =& \text{tr}\left(\mathbb{E}\left[\left(Q_k^L - Q^*\right)\left(Q_k^L - Q^*\right)^T\right]\right) \\
    =& \text{tr}(X_k)
\end{align*}
and~\cref{lem3.12}, we have
\begin{align*}
    & \mathbb{E}\left[\left\|Q_k^L - Q^*\right\|_2^2\right] \\
    \le& \frac{6\alpha \left(\ln (|\mathcal{A}|)+\beta\right)^2|\mathcal{S} \times \mathcal{A}|^2}{\beta^2 d_{min}(1-\gamma)^3} + |\mathcal{S} \times \mathcal{A}|^2 \left\|Q_0^L - {Q^*}\right\|_2^2 \rho^{2k}.
\end{align*}

Taking the square root on both sides of the above inequality and combining the relation $\mathbb{E}\left[\left\| \cdot \right\|_2\right] = \mathbb{E}[\sqrt{\| \cdot \|_2^2}] \le \sqrt{\mathbb{E}[\| \cdot \|_2^2]} $ leads to
\begin{align*}
    & \mathbb{E}\left[\left\|Q_k^L - Q^*\right\|_2\right] \\
    \le& \sqrt{\frac{6\alpha \left(\ln (|\mathcal{A}|)+\beta\right)^2|\mathcal{S} \times \mathcal{A}|^2}{\beta^2 d_{min}(1-\gamma)^3} + |\mathcal{S} \times \mathcal{A}|^2 \left\|Q_0^L - {Q^*}\right\|_2^2 \rho^{2k}} \\     
    \le& \frac{\sqrt{6} \alpha^\frac{1}{2}\left(\ln (|\mathcal{A}|)+\beta\right) |\mathcal{S} \times \mathcal{A}|}{\beta d_{min}^\frac{1}{2}(1-\gamma)^\frac{3}{2}} + |\mathcal{S} \times \mathcal{A}| \left\|Q_0^L - {Q^*}\right\|_2 \rho^k,
\end{align*}
where the subadditivity of the square root function is utilized in the last inequality.
\end{proof} \vspace{0.3cm}

Note that the first term on the right-hand side of~\eqref{eq: final iteration lower} is the constant error that can be minimized by the smaller step size $\alpha \in (0,1)$ and the larger parameter $\beta$. The second term $O(\rho^k)$ exponentially decays as $k \rightarrow \infty$.

\subsection{Upper Comparison System of LSE Soft Q-learning}
Let us consider the affine system
\begin{equation}\label{eq: soft q-learning upper comparison system}
\begin{aligned}
    Q_{k+1}^U-Q^* =& A_{Q_k^{LSE}}\left(Q_k^U-Q^*\right) \\
    &+ \alpha w_k^{LSE} + \alpha\gamma DP \frac{\ln(|\mathcal{A}|)}{\beta} \textbf{1}
\end{aligned}
\end{equation}
with
\[A_{Q_k^{LSE}} := I+\alpha\left(\gamma DP\Pi _{Q_k^{LSE}}^{\max}-D\right)\]
and the stochastic noise~\eqref{eq: lse noise}, which is called \textit{upper comparison system of LSE soft Q-learning}. Supported by the following proposition, this system can be an upper bound for the LSE soft Q-learning iteration.

\vspace{0.3cm}
\begin{proposition} \label{prop3.4}
    Suppose $Q_0^U-Q^* \ge Q_0^{LSE}-Q^*$, where $Q_0^U-Q^* \in \mathbb{R}^{|\mathcal{S}||\mathcal{A}|}$ and ``$\ge$" is the element-wise inequality. Then, for all $k \ge 0$, 
    \[Q_k^U-Q^* \ge Q_k^{LSE}-Q^*.\]
\end{proposition} \vspace{0.3cm}

\begin{proof}
To apply the induction argument, assume that $Q_k^U-Q^* \ge Q_k^{LSE}-Q^*$ holds for some $k \ge 0$. Then, we have
\begin{align*}
    & Q_{k+1}^{LSE}-Q^* \\
    =& Q_k^{LSE} - {Q^*} \\
    &+ \alpha \left\{ DR + \gamma DP H_{LSE}^{\beta}\left(Q_k^{LSE}\right) - DQ_k^{LSE} + w_k^{LSE}\right\}\\   
    \le& Q_k^{LSE} - {Q^*} + \alpha DR \\
    &+ \alpha\gamma DP \left({\Pi _{Q_k^{LSE}}^{\max}}Q_k^{LSE} + \frac{\ln(|\mathcal{A}|)}{\beta} \textbf{1} \right) \\
    &+ \alpha \left(- DQ_k^{LSE} + w_k^{LSE} \right)\\ 
    =& Q_k^{LSE} - {Q^*} - \alpha \left(\gamma DP{\Pi _{Q^*}^{\max}} - D\right)Q^* \\
    &+ \alpha\gamma DP \left({\Pi _{Q_k^{LSE}}^{\max}}Q_k^{LSE} + \frac{\ln(|\mathcal{A}|)}{\beta} \textbf{1} \right)\\
    &+ \alpha\gamma DP\left(\Pi _{Q_k^{LSE}}^{\max}Q^* - \Pi _{Q_k^{LSE}}^{\max}Q^* \right) \\
    &+ \alpha \left(- DQ_k^{LSE} + w_k^{LSE}\right) \\
    =& \left\{I+ \alpha \left(\gamma DP{\Pi _{Q_k^{LSE}}^{\max}} - D\right)\right\}\left(Q_k^{LSE}-Q^*\right) \\
    &+ \alpha \gamma DP \left(\Pi _{Q_k^{LSE}}^{\max}-\Pi _{Q^*}^{\max}\right)Q^*\\
    &+ \alpha w_k^{LSE} + \alpha\gamma DP \frac{\ln(|\mathcal{A}|)}{\beta} \textbf{1} \\
    \le& \left\{I+ \alpha \left(\gamma DP{\Pi _{Q_k^{LSE}}^{\max}} - D\right)\right\}\left(Q_k^{LSE}-Q^*\right) \\
    &+ \alpha w_k^{LSE} + \alpha\gamma DP \frac{\ln(|\mathcal{A}|)}{\beta} \textbf{1} \\
    \le& \left\{I+ \alpha \left(\gamma DP{\Pi _{Q_k^{LSE}}^{\max}} - D\right)\right\}\left(Q_k^U - {Q^*}\right) \\
    &+ \alpha w_k^{LSE} + \alpha\gamma DP \frac{\ln(|\mathcal{A}|)}{\beta} \textbf{1} \\
    =& Q_{k+1}^U-Q^*,
\end{align*}
where~\cref{prop3.1} is used in the first inequality and the optimal Bellman equation is utilized in the second equality. The second inequality holds because $\alpha \gamma DP\left(\Pi _{Q_k^{LSE}}^{\max}-\Pi _{Q^*}^{\max}\right)Q^* \le $$ \alpha \gamma DP \left(\Pi _{Q^*}^{\max}-\Pi _{Q^*}^{\max} \right)Q^*=0$. Moreover, the last inequality relies on the hypothesis $Q_k^U-Q^* \ge Q_k^{LSE}-Q^*$ and the fact that all elements of $I+ \alpha \left( \gamma DP{\Pi _{Q_k^{LSE}}^{\max}} - D \right)$ are non-negative. This brings the proof to its conclusion.
\end{proof} \vspace{0.3cm}

The upper comparison system~\eqref{eq: soft q-learning upper comparison system} includes the non-constant matrix $A_{Q_k^{LSE}}$ rather than $A_{Q^*}$, and $A_{Q_k^{LSE}}$ switches according to the variations of $Q_k^{LSE}$. This can be a problem for analyzing the stability since $A_{Q_k^{LSE}}$ and $Q_k^U-Q^*$ depend on the same stochastic noise~\eqref{eq: lse noise}, and hence, they are statistically dependent. This means that the linear recursion in~\eqref{autocorrelation matrix} cannot be used again for our analysis of the upper comparison system. To overcome this obstacle, we will apply a different strategy for the analysis of the upper comparison system, which is presented in the next subsection.

\subsection{Error System for LSE Soft Q-learning System Analysis}
To overcome the aforementioned difficulty, we examine an error system created by subtracting the lower comparison system~\eqref{eq: soft q-learning lower comparison system} from the upper comparison system~\eqref{eq: soft q-learning upper comparison system}:
\begin{equation}\label{eq: subtracted upper comparison system}
\begin{aligned}
    & Q_{k+1}^U-Q_{k+1}^L \\
    =& A_{Q_k^{LSE}}\left(Q_k^U-Q_k^L\right) \\
    &+ B_{Q_k^{LSE}}\left(Q_k^L-Q^*\right) + \alpha\gamma DP \frac{\ln(|\mathcal{A}|)}{\beta} \textbf{1}
\end{aligned}
\end{equation}
with
\[B_{Q_k^{LSE}} := A_{Q_k^{LSE}}-A_{Q^*}.\]

In this case, the stochastic noise $w_k^{LSE}$ is eliminated from the error system of LSE soft Q-learning. Here, $A_{Q_k^{LSE}}$ and $B_{Q_k^{LSE}}$ alter in response to $Q_k^{LSE}$, and $Q_k^L-Q*$ can be regarded as an external disturbance.

The subsequent analysis is based on the following intuition: as $Q_k^L \rightarrow Q^*$ in~\cref{thm3.1}, we can easily show $Q_k^{LSE} \rightarrow Q^*$ if we can prove $Q_k^U-Q_k^L \rightarrow 0$ as $k \rightarrow \infty$ since
\begin{equation} \label{eq: Q_k-Q^* inequality}
\begin{aligned}
    & \mathbb{E}\left[\left\|Q_k^{LSE}-Q^*\right\|_{\infty}\right] \\
    =& \mathbb{E}\left[\left\|Q_k^{LSE}-Q_k^L+Q_k^L-Q^*\right\|_{\infty}\right] \\
    \le& \mathbb{E}\left[\left\|Q_k^{LSE}-Q_k^L\right\|_{\infty}\right]+ \mathbb{E}\left[\left\|Q_k^L-Q^*\right\|_{\infty}\right] \\
    \le& \mathbb{E}\left[\left\|Q_k^U-Q_k^L\right\|_{\infty}\right]+ \mathbb{E}\left[\left\|Q_k^L-Q^*\right\|_{\infty}\right],
\end{aligned}
\end{equation}
where the last inequality is derived from~\cref{prop3.4}.

Then, we get a finite-time error bound for the LSE soft Q-learning in the following theorem.
\vspace{0.3cm}
\begin{theorem} \label{thm3.2}
For any $k \ge 0$, we have
\begin{equation} \label{eq: final iteration upper}
\begin{aligned}
    & \mathbb{E}\left[\left\|Q_k^{LSE}-Q^*\right\|_{\infty}\right] \\
    \le& \frac{3\sqrt{6} \alpha^\frac{1}{2} d_{max} \left(\ln (|\mathcal{A}|)+\beta\right) |\mathcal{S} \times \mathcal{A}|}{\beta d_{min}^\frac{3}{2}(1-\gamma)^\frac{5}{2}} \\
    &+ \frac{4\alpha\gamma d_{max} |\mathcal{S} \times \mathcal{A}|^\frac{3}{2}}{1-\gamma} k \rho^{k-1} + \frac{\ln(|\mathcal{A}|)}{{\beta {d_{\min }}(1 - \gamma )}} \\
    &+ \frac{2|\mathcal{S} \times \mathcal{A}|^\frac{3}{2}}{1-\gamma}\rho^k.
\end{aligned}
\end{equation}
\end{theorem} \vspace{0.3cm}

\begin{proof}
Considering the norm of~\eqref{eq: subtracted upper comparison system}, we obtain
\begin{align}
    & \left\|Q_{i+1}^U-Q_{i+1}^L\right\|_{\infty} \nonumber \\
    \le& \left\|A_{Q_i^{LSE}}\right\|_{\infty} \left\|Q_i^U-Q_i^L\right\|_{\infty} + \left\|B_{Q_i^{LSE}}\right\|_{\infty} \left\|Q_i^L-Q^*\right\|_{\infty} \nonumber \\
    &+ \left\|\alpha\gamma DP \frac{\ln(|\mathcal{A}|)}{\beta} \textbf{1} \right\|_{\infty} \nonumber \\
    \le& \rho \left\|Q_i^U-Q_i^L\right\|_{\infty} \nonumber \\
    &+ 2\alpha\gamma d_{max} \left\|Q_i^L-Q^*\right\|_{\infty} + \frac {\alpha}{\beta} \ln(|\mathcal{A}|) \label{lse upper minus lower}
\end{align}
using~\cref{lem3.2} and the following results:
\begin{align*}
    \|B_{Q_i^{LSE}}\|_{\infty} =& \|A_{Q_i^{LSE}}-A_{Q^*}\|_{\infty} \\
    \le& \alpha \gamma d_{max} \left\|P \left(\Pi_{Q_i^{LSE}}^{\max} - \Pi _{Q^*}^{\max} \right) \right\|_{\infty} \\
    \le& 2\alpha \gamma d_{max}
\end{align*}
and
\begin{align*}
    \left\| \alpha\gamma DP \frac{\ln(|\mathcal{A}|)}{\beta} \textbf{1} \right\|_{\infty} 
    \le& \alpha \gamma \|D\|_{\infty} \|P\|_{\infty} \left\|\frac {\ln(|\mathcal{A}|)}{\beta} \textbf{1}\right\|_{\infty} \\
    \le& \alpha \gamma d_{max} \frac {\ln(|\mathcal{A}|)}{\beta} \\
    \le& \frac {\alpha}{\beta} \ln(|\mathcal{A}|).
\end{align*}

Taking the expectation of~\eqref{lse upper minus lower} and applying~\cref{thm3.1} leads to
\begin{align*}
    & \mathbb{E} \left[ \left\|Q_{i+1}^U-Q_{i+1}^L\right\|_{\infty} \right] \\
    \le& \rho\mathbb{E} \left[ \left\|Q_i^U-Q_i^L \right\|_{\infty} \right] + 2\alpha\gamma d_{max}\mathbb{E} \left[ \left\|Q_i^L-Q^* \right\|_2 \right] \\
    &+ \frac {\alpha}{\beta} \ln(|\mathcal{A}|) \\
    \le& \rho E\left[ \left\|Q_i^U-Q_i^L \right\|_{\infty} \right] \\
    &+ \frac{2\sqrt{6} \alpha^\frac{3}{2} \gamma d_{max} \left(\ln (|\mathcal{A}|)+\beta \right) |\mathcal{S} \times \mathcal{A}|}{\beta d_{min}^\frac{1}{2}(1-\gamma)^\frac{3}{2}} \\
    &+ 2\alpha\gamma d_{max}|\mathcal{S} \times \mathcal{A}| \left\|Q_0^L-Q^* \right\|_2\rho^i + \frac {\alpha}{\beta} \ln(|\mathcal{A}|).
\end{align*}
  
Then, applying the inequality recursively and letting $Q_0^U=Q_0^L$ produce
\begin{align*}
    & \mathbb{E} \left[ \left\|Q_k^U-Q_k^L \right\|_{\infty} \right] \\
    \le& \frac{2\sqrt{6} \alpha^\frac{3}{2} \gamma d_{max} \left(\ln (|\mathcal{A}|)+\beta \right) |\mathcal{S} \times \mathcal{A}|}{\beta d_{min}^\frac{1}{2}(1-\gamma)^\frac{3}{2}} \sum_{i=0}^{k-1} \rho^{i} \\
    &+ 2\alpha\gamma d_{max}|\mathcal{S} \times \mathcal{A}| \left\|Q_0^L-Q^*\right\|_2 k \rho^{k-1} \\
    &+ \frac{\alpha\ln(|\mathcal{A}|)}{\beta} \sum_{i=0}^{k-1} \rho^{i}.
\end{align*}

Noting the relation
\begin{equation} \label{Q_0^L-Q^* inequality}
\begin{aligned}
    \left\|Q_0^L-Q^*\right\|_2 \le& |\mathcal{S} \times \mathcal{A}|^\frac{1}{2} \left\|Q_0^L-Q^*\right\|_{\infty} \\
    \le& |\mathcal{S} \times \mathcal{A}|^\frac{1}{2} \left( \left\|Q_0^L\right\|_{\infty} + \left\|Q^*\right\|_{\infty} \right) \\
    \le& |\mathcal{S} \times \mathcal{A}|^\frac{1}{2} \left(1+\frac{1}{1-\gamma} \right) \\
    \le& |\mathcal{S} \times \mathcal{A}|^\frac{1}{2} \frac{2}{1-\gamma}
\end{aligned}
\end{equation}
and $\sum_{i=0}^{k-1} \rho^{i} \le \sum\limits_{i = 0}^\infty  {{\rho ^i}} \le \frac{1}{1 - \rho}$ with~\cref{as: 1,lem3.6,def3}, the last inequality turns into
\begin{align*}
    \mathbb{E} \left[ \left\|Q_k^U-Q_k^L\right\|_{\infty} \right] \le& \frac{2\sqrt{6} \alpha^\frac{1}{2} \gamma d_{max} \left(\ln (|\mathcal{A}|)+\beta \right) |\mathcal{S} \times \mathcal{A}|}{\beta d_{min}^\frac{3}{2}(1-\gamma)^\frac{5}{2}} \\
    &+ \frac{4\alpha\gamma d_{max} |\mathcal{S} \times \mathcal{A}|^\frac{3}{2}}{1-\gamma} k \rho^{k-1} \\
    &+ \frac{\ln(|\mathcal{A}|)}{{\beta {d_{\min }}(1 - \gamma )}}.
\end{align*}

Subsequently, combining~\eqref{eq: Q_k-Q^* inequality},~\cref{thm3.1}, and the above inequality leads to the desired result as follows:
\begin{align*}
    & \mathbb{E} \left[ \left\|Q_k^{LSE}-Q^*\right\|_{\infty} \right] \\
    \le& \mathbb{E} \left[ \left\|Q_k^U-Q_k^L\right\|_{\infty} \right]+ \mathbb{E} \left[ \left\|Q_k^L-Q^*\right\|_{\infty} \right] \\
    \le& \frac{2\sqrt{6} \alpha^\frac{1}{2} \gamma d_{max} \left(\ln (|\mathcal{A}|)+\beta \right) |\mathcal{S} \times \mathcal{A}|}{\beta d_{min}^\frac{3}{2}(1-\gamma)^\frac{5}{2}} \\
    &+ \frac{4\alpha\gamma d_{max} |\mathcal{S} \times \mathcal{A}|^\frac{3}{2}}{1-\gamma} k \rho^{k-1} + \frac{\ln(|\mathcal{A}|)}{{\beta {d_{\min }}(1 - \gamma )}} \\
    &+ \frac{\sqrt{6} \alpha^\frac{1}{2} \left(\ln (|\mathcal{A}|)+\beta \right) |\mathcal{S} \times \mathcal{A}|}{\beta d_{min}^\frac{1}{2}(1-\gamma)^\frac{3}{2}} \\
    &+ |\mathcal{S} \times \mathcal{A}| \left\|Q_0^L-Q^*\right\|_2\rho^k \\
    \le& \frac{2\sqrt{6} \alpha^\frac{1}{2} \gamma d_{max} \left(\ln (|\mathcal{A}|)+\beta \right) |\mathcal{S} \times \mathcal{A}|}{\beta d_{min}^\frac{3}{2}(1-\gamma)^\frac{5}{2}} \\
    &+ \frac{4\alpha\gamma d_{max} |\mathcal{S} \times \mathcal{A}|^\frac{3}{2}}{1-\gamma} k \rho^{k-1} + \frac{\ln(|\mathcal{A}|)}{{\beta {d_{\min }}(1 - \gamma )}} \\
    &+ \frac{\sqrt{6} \alpha^\frac{1}{2} \left(\ln (|\mathcal{A}|)+\beta \right) |\mathcal{S} \times \mathcal{A}|}{\beta d_{min}^\frac{1}{2}(1-\gamma)^\frac{3}{2}} + \frac{2|\mathcal{S} \times \mathcal{A}|^\frac{3}{2}}{1-\gamma}\rho^k \\
    \le& \frac{3\sqrt{6} \alpha^\frac{1}{2} d_{max} \left(\ln (|\mathcal{A}|)+\beta \right) |\mathcal{S} \times \mathcal{A}|}{\beta d_{min}^\frac{3}{2}(1-\gamma)^\frac{5}{2}} \\
    &+ \frac{4\alpha\gamma d_{max} |\mathcal{S} \times \mathcal{A}|^\frac{3}{2}}{1-\gamma} k \rho^{k-1} + \frac{\ln(|\mathcal{A}|)}{{\beta {d_{\min }}(1 - \gamma )}} \\
    &+ \frac{2|\mathcal{S} \times \mathcal{A}|^\frac{3}{2}}{1-\gamma}\rho^k,
\end{align*}
where~\eqref{Q_0^L-Q^* inequality} is used in the third inequality and $\gamma \in [0,1)$,~\cref{def1} is utilized in the last inequality.
\end{proof} \vspace{0.3cm}

On the right-hand side of~\eqref{eq: final iteration upper}, the second and third terms result from the difference between the upper comparison system and the lower system, while the last term results from the difference between the lower comparison system and the original system. Furthermore, the second term $O(k\rho^{k-1})$ and the last term $O(\rho^k)$ exponentially decay as $k \rightarrow \infty$. The first and third error terms can be reduced by decreasing the step size $\alpha$ and increasing the parameter $\beta$, respectively.

\section{FINITE-TIME ANALYSIS OF BOLTZMANN SOFT Q-LEARNING}
In this section, we study the finite-time error analysis of Boltzmann soft Q-learning. Boltzmann soft Q-learning can be demonstrated in a similar manner to the previous one. However, as a result of changing the bounds for the operator, an alternative analysis in place of the autocorrelation matrix is required to establish the finite-time error bound of the lower comparison system.

\subsection{Nonlinear System Representation of Boltzmann Soft Q-learning}
Boltzmann soft Q-learning can be expressed as
\begin{equation*}
\begin{aligned}
    Q_{k + 1}^{Boltz} =& Q_k^{Boltz} + \alpha \Big\{ DR + \gamma DP H_{Boltz}^{\beta}\left(Q_k^{Boltz}\right) \\
    &- DQ_k^{Boltz} + w_k^{Boltz} \Big\},
    \end{aligned}
\end{equation*}
where
\[ H_{Boltz}^{\beta}\left(Q_k\right)
:= \left[ {\begin{array}{*{20}{c}}
h_{Boltz}^{\beta}(Q_k(1,\cdot)) \\
 \vdots \\
h_{Boltz}^{\beta}(Q_k(|\mathcal{S}|,\cdot))
\end{array}} \right],\]
\begin{equation} \label{eq: boltz noise}
\begin{aligned}
    & w_k^{Boltz} \\
    =& (e_{a_k} \otimes e_{s_k})r_k \\
    &+ \gamma(e_{a_k} \otimes e_{s_k})(e_{s_k'})^T  H_{Boltz}^{\beta}\left(Q_k^{Boltz}\right) \\
    &- (e_{a_k} \otimes e_{s_k})(e_{a_k} \otimes e_{s_k})^T Q_k^{Boltz} \\
    &- \Big\{DR+\gamma DP H_{Boltz}^{\beta}\left(Q_k^{Boltz}\right) - DQ_k^{Boltz}\Big\}.
\end{aligned}
\end{equation}

This nonlinear system can also be examined more easily with comparison systems, which are constructed on the basis of the following proposition:
\vspace{0.3cm}
\begin{proposition} \label{prop4}
For any $Q \in {\mathbb R}^{|\mathcal{S}||\mathcal{A}|}$, we have
\[\max_{a \in \mathcal{A}}Q(s,a) - \frac{\ln(|\mathcal{A}|)}{\beta} \le h_{Boltz}^{\beta}(Q(s,\cdot)) \le \max_{a \in \mathcal{A}}Q(s,a)\]
and hence
\[{\Pi_Q^{\max}}Q - \frac{{\ln (|\mathcal{A}|)}}{\beta }{\bf{1}} \le H_{Boltz}^{\beta}\left(Q\right) \le {\Pi_Q^{\max}}Q,\]
where ``$\le$'' denotes the element-wise inequality and $\textbf{1}$ is a column vector with all elements equal to 1.
\end{proposition} \vspace{0.3cm}
The proof of~\cref{prop4} is supplied in Appendix.

\subsection{Lower Comparison System of Boltzmann Soft Q-learning}
We take the lower comparison system of Boltzmann soft Q-learning as
\begin{equation}\label{eq: Boltzmann soft q-learning lower comparison system}
\begin{aligned}
    Q_{k+1}^L-Q^* =& A_{Q^*}\left(Q_k^L-Q^*\right) \\
    &+ \alpha w_k^{Boltz} - \alpha \gamma DP \frac{\ln(|\mathcal{A}|)}{\beta} \textbf{1}
\end{aligned}
\end{equation}
with the stochastic noise~\eqref{eq: boltz noise}, which can be proved with the following proposition. The proof is included in Appendix.

\vspace{0.3cm}
\begin{proposition} \label{prop lower boltz}
    Suppose $Q_0^L - {Q^*} \le Q_0^{Boltz} - {Q^*}$, where $Q_0^L-Q^* \in \mathbb{R}^{|\mathcal{S}||\mathcal{A}|}$ and ``$\le$" is the element-wise inequality. Then, for all $k \ge 0$, 
    \[Q_k^L - {Q^*} \le Q_k^{Boltz} - {Q^*}.\]
\end{proposition} \vspace{0.3cm}

Here, the extra term $\alpha \gamma DP \frac{\ln(|\mathcal{A}|)}{\beta} \textbf{1}$ in~\eqref{eq: Boltzmann soft q-learning lower comparison system} makes it challenging to create the autocorrelation matrix in the form of~\eqref{autocorrelation matrix}. Therefore, we generate a finite-time error bound for the lower comparison system of Boltzmann soft Q-learning directly, leading to the following theorem:
\vspace{0.3cm}
\begin{theorem}\label{thm3.3}
For any $k \ge 0$, we have
\begin{equation}\label{eq: Boltzmann final iteration lower}
\begin{aligned}
    \mathbb{E}\left[\left\|Q_k^L - Q^*\right\|_2\right] &\le |\mathcal{S} \times \mathcal{A}|^\frac{1}{2} \left\|Q_0^L-Q^*\right\|_2 \rho^{k} \\
    &+ \frac{\sqrt{6} \alpha^\frac{1}{2} \left(\ln (|\mathcal{A}|)+\beta\right) |\mathcal{S} \times \mathcal{A}|^\frac{1}{2}}{\beta d_{min}^\frac{1}{2}(1-\gamma)^\frac{3}{2}} \\
    &+ \frac {\gamma d_{max} \ln(|\mathcal{A}|)|\mathcal{S} \times \mathcal{A}|^\frac{1}{2}}{\beta d_{min}(1-\gamma)}.
\end{aligned}
\end{equation}
\end{theorem} \vspace{0.3cm}

The proof of~\cref{thm3.3} is included in Appendix.

The first term $O(\rho^k)$ on the right-hand side of~\eqref{eq: Boltzmann final iteration lower} exponentially vanishes when $k \rightarrow \infty$. Moreover, we can reduce the constant error, the second and last terms of~\eqref{eq: Boltzmann final iteration lower}, by decreasing the step size $\alpha$ and increasing the parameter $\beta$.

\subsection{Upper Comparison System of Boltzmann Soft Q-learning}
We get the upper comparison system of Boltzmann soft Q-learning as
\begin{equation} \label{eq: Boltzmann upper comparison system}
    Q_{k+1}^U-Q^* = A_{Q_k^{Boltz}}\left(Q_k^U-Q^*\right) + \alpha w_k^{Boltz}
\end{equation}
with
\[A_{Q_k^{Boltz}} := I+\alpha \left(\gamma DP\Pi _{Q_k^{Boltz}}^{\max}-D \right)\]
and the stochastic noise~\eqref{eq: boltz noise}, which can be proved with the following proposition. The proof is available in Appendix.

\vspace{0.3cm}
\begin{proposition} \label{prop upper boltz}
    Suppose $Q_0^U-Q^* \ge Q_0^{Boltz}-Q^*$, where $Q_0^U-Q^* \in \mathbb{R}^{|\mathcal{S}||\mathcal{A}|}$ and ``$\ge$" is the element-wise inequality. Then, for all $k \ge 0$, 
    \[Q_k^U-Q^* \ge Q_k^{Boltz}-Q^*.\]
\end{proposition} \vspace{0.3cm}

\subsection{Error System for Boltzmann Soft Q-learning System Analysis}
To circumvent the issue of statistical dependence in~\eqref{eq: Boltzmann upper comparison system}, we can obtain the error system as
\begin{equation} \label{eq: Boltzmann subtracted upper comparison system}
\begin{aligned}
    & Q_{k+1}^U-Q_{k+1}^L \\
    =& A_{Q_k^{Boltz}}\left(Q_k^U-Q_k^L\right) \\
    &+ B_{Q_k^{Boltz}}\left(Q_k^L-Q^*\right) + \alpha \gamma DP \frac{\ln(|\mathcal{\mathcal{A}}|)}{\beta} \textbf{1}
\end{aligned}
\end{equation}
with
\[B_{Q_k^{Boltz}} := A_{Q_k^{Boltz}}-A_{Q^*}\]
by subtracting the lower comparison system~\eqref{eq: Boltzmann soft q-learning lower comparison system} from the upper comparison system~\eqref{eq: Boltzmann upper comparison system}. In this case, we find that~\eqref{eq: Boltzmann subtracted upper comparison system} and~\eqref{eq: subtracted upper comparison system} are practically the same. Hence, we can get a finite-time error bound for Boltzmann soft Q-learning in the same vein with~\cref{thm3.2}.

\vspace{0.3cm}
\begin{theorem} \label{boltzmann final upper}
For any $k \ge 0$, we have
\begin{equation} \label{eq: boltzmann final iteration upper}
\begin{aligned}
    & \mathbb{E}\left[\left\|Q_k^{Boltz}-Q^*\right\|_{\infty}\right] \\
    \le& \frac{4\alpha\gamma d_{max} |\mathcal{S} \times \mathcal{A}|}{1-\gamma} k \rho^{k-1} \\
    &+ \frac{3\sqrt{6} \alpha^\frac{1}{2} d_{max} \left(\ln (|\mathcal{A}|)+\beta\right) |\mathcal{S} \times \mathcal{A}|^\frac{1}{2}}{\beta d_{min}^\frac{3}{2}(1-\gamma)^\frac{5}{2}} \\
    &+ \frac {4 d_{max} \ln(|\mathcal{A}|) |\mathcal{S} \times \mathcal{A}|^\frac{1}{2}}{\beta d_{min}^2(1-\gamma)^2} + \frac{2 |\mathcal{S} \times \mathcal{A}|}{1-\gamma} \rho^k.
\end{aligned}
\end{equation}
\end{theorem} \vspace{0.3cm}

The proof of~\cref{boltzmann final upper} is given in Appendix.

On the right-hand side of~\eqref{eq: boltzmann final iteration upper}, the first term results from the difference between the upper comparison system and the lower system, while the last term results from the difference between the lower comparison system and the original system. Furthermore, the first term $O(k\rho^{k-1})$ and the last term $O(\rho^k)$ exponentially decay as $k \rightarrow \infty$. The second and third error terms can be reduced by the smaller step size $\alpha$ and the larger parameter $\beta$, respectively.

\section{EMPIRICAL RESULTS}

\begin{figure}[ht]
    \centering
    \includegraphics[width=0.9\linewidth]{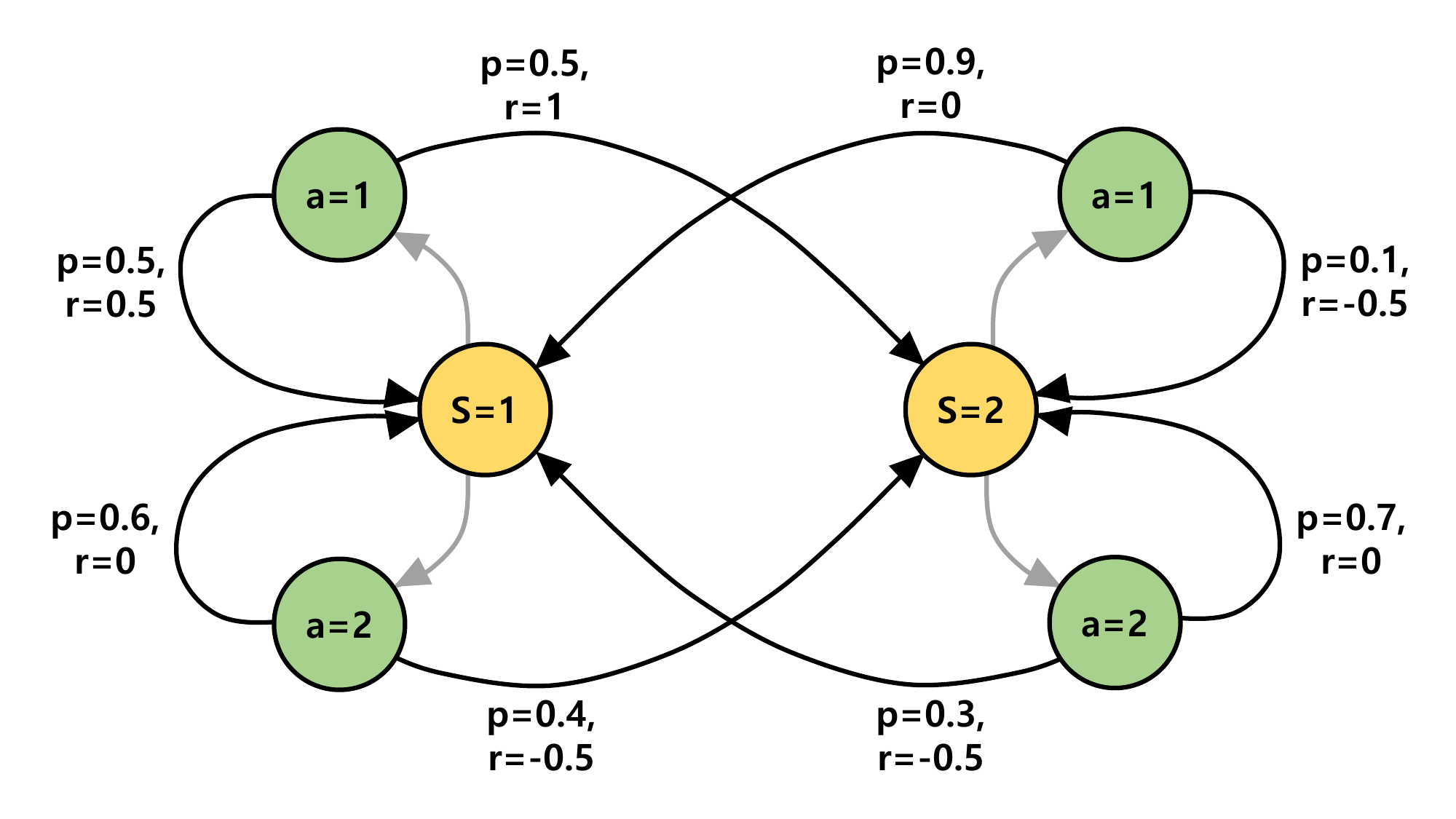}
    \caption{We consider MDP that has $\mathcal{S}=\{1,2\}$, $\mathcal{A}=\{1,2\}$, state transition matrix for each action $p_1=[[0.5, 0.5], [0.9, 0.1]]$ and $p_2=[[0.6, 0.4], [0.3, 0.7]]$, and the reward function $r(1,1,1)=0.5$, $r(1,1,2)=1$, $r(1,2,2)=r(2,1,2)=r(2,2,1)=-0.5$ and $0$ in other cases.}
    \label{fig1}
\end{figure}

In order to verify the validity of our analysis, we consider an MDP presented in~\cref{fig1}. This MDP has $\mathcal{S}=\{1,2\}$, $\mathcal{A}=\{1,2\}$, $\gamma=0.9$, and the state transition matrix for each action $p_1=[[0.5, 0.5], [0.9, 0.1]]$ and $p_2=[[0.6, 0.4], [0.3, 0.7]]$. We set the reward function to be $r(1,1,1)=0.5$, $r(1,1,2)=1$, $r(1,2,2)=r(2,1,2)=r(2,2,1)=-0.5$ and $0$ in other cases. The initial state is chosen along the initial state distribution $[0.8, 0.2]$, and the episode terminates if the maximum number of steps 50 is achieved. We use the softmax policy as the behavior policy for each soft Q-learning algorithm: $\pi_b(a|s) = \frac{\exp(Q(s,a))}{\sum_{u \in \mathcal{A}} \exp(Q(s,u))}$.

\begin{figure}[ht]
    \centering
    \begin{subfigure}{\columnwidth}
    \includegraphics[width=\columnwidth]{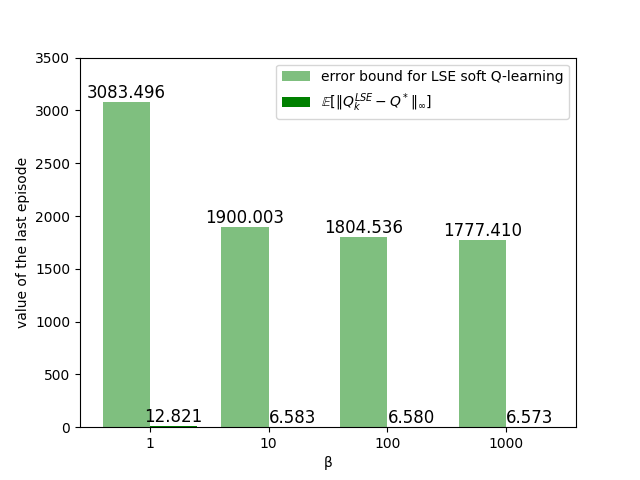} 
    \caption{LSE soft Q-learning}
    \label{fig2a}
    \end{subfigure}
    \begin{subfigure}{\columnwidth}
    \includegraphics[width=\columnwidth]{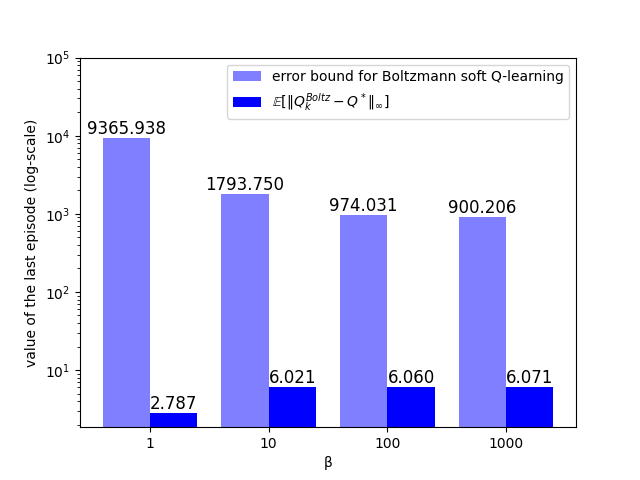}
    \caption{Boltzmann soft Q-learning}
    \label{fig2b}
    \end{subfigure}

    \caption{Impact of $\beta$ on $\mathbb{E}[\|Q_{\infty}-Q^*\|_\infty]$ and the finite-time error bounds when $\alpha=0.001$ }
    \label{fig2}
\end{figure}

\begin{figure}[ht]
    \centering
    \begin{subfigure}{\columnwidth}
    \includegraphics[width=\columnwidth]{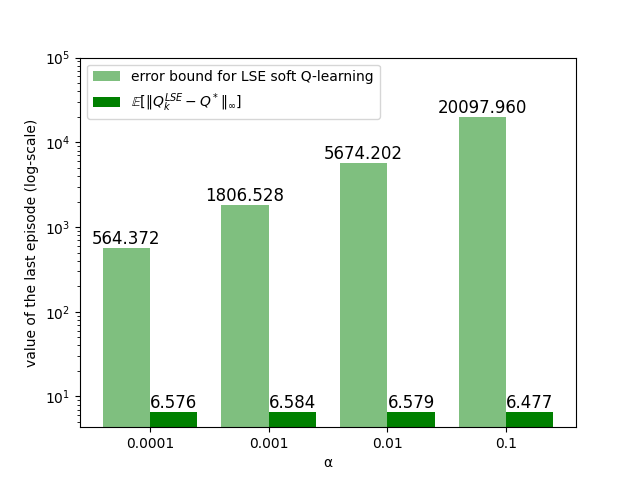} 
    \caption{LSE soft Q-learning}
    \label{fig3a}
    \end{subfigure}
    \begin{subfigure}{\columnwidth}
    \includegraphics[width=\columnwidth]{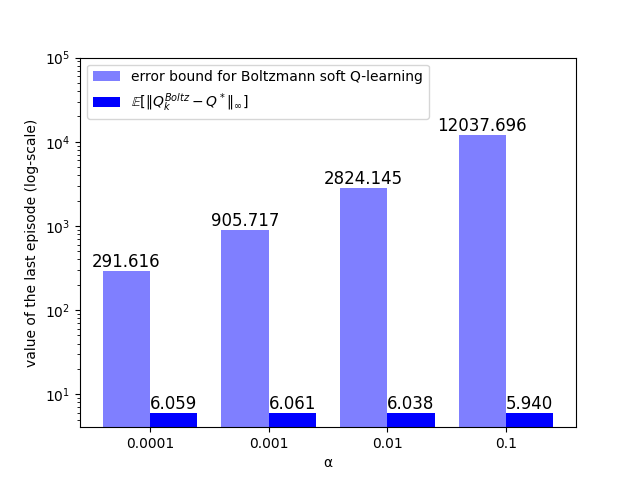}
    \caption{Boltzmann soft Q-learning}
    \label{fig3b}
    \end{subfigure}
    
    \caption{Impact of $\alpha$ on $\mathbb{E}[\|Q_{\infty}-Q^*\|_\infty]$ and the finite-time error bounds when $\beta=1000$ }
    \label{fig3}
\end{figure}

\cref{fig2} and~\cref{fig3} illustrate the infinity norm of the errors between the fixed point of each algorithm and the optimal Q-function, $\mathbb{E}[\|Q_{\infty}-Q^*\|_\infty]$, together with their finite-time error bounds in the right-hand side of~\cref{thm3.2,boltzmann final upper} at the last episode. In particular,~\cref{fig2} shows how the value of $\beta$ affects~\cref{thm3.2,boltzmann final upper} with $\alpha=0.001$, while~\cref{fig3} shows how the value of $\alpha$ influences when $\beta=1000$. Note that although the extreme values of $\alpha$ and $\beta$ might provide dramatic outcomes, we select the values within the suitable range that is usually used in practice. Furthermore, we plot each graph using an average over 10 runs. The fixed point is obtained after $10^5$ iterations in a single run.

In~\cref{fig2}, the LSE soft Q-learning case is depicted in~\cref{fig2a}, while the Boltzmann soft Q-learning case is presented in~\cref{fig2b}. As the parameter $\beta$ grows in both algorithms, the finite-time error bounds decrease. Note that as beta grows, the rate of the finite-time error bound change falls, since beta appears in both the numerator and denominator of the first term of~\cref{thm3.2} and the second term of~\cref{boltzmann final upper}, respectively. Furthermore, in~\cref{fig2a}, $\mathbb{E}[\|Q-Q^*\|_\infty]$ drops because the LSE operator approximates the max operator with the higher $\beta$ and has a unique fixed point. However, in~\cref{fig2b}, $\mathbb{E}[\|Q-Q^*\|_\infty]$ doesn't reduce as $\beta$ increases since the Boltzmann operator is not a non-expansion, hence the fixed point can be changed by the parameter $\beta$~\cite{asadi2017alternative}. In this scenario, our methodology can serve as a preliminary and simplified evaluation of the convergence prior to performing a more complex finite-time analysis. 

In~\cref{fig3}, the LSE soft Q-learning case is represented in~\cref{fig3a}, while the Boltzmann soft Q-learning case is provided in~\cref{fig3b}. As $\alpha$ lowers in both algorithms, the finite-time error bounds drop as well. Moreover, since the value of $\beta$ is fixed, $\mathbb{E}[\|Q-Q^*\|_\infty]$ values of each algorithm are almost identical.

Based on these results, we can conclude that our analysis can be utilized to estimate the behavior of LSE and Boltzmann soft Q-learning. Moreover, we can obtain a tighter bound for each soft Q-learning with higher $\beta$ and smaller $\alpha$ values.

\section{CONCLUSION}
In this paper, we present a unique finite-time error analysis for soft Q-learning algorithms that use LSE and Boltzmann operators. Specifically, using the switching system approach, we find the upper and lower bounds of two operators and offer the upper and lower comparison systems of soft Q-learning. By proving that each comparison system converges, it is simple to provide a finite-time analysis of soft Q-learning. We believe that the suggested approach can give further perspectives using simple control-theoretic concepts and supplement the study on the analysis of soft Q-learning. For future works, obtaining tighter error bounds will be beneficial for the next steps. Moreover, we believe that the proposed approach will possibly function as a unified analysis framework to handle further soft Q-learning variations.

\bibliographystyle{IEEEtran}
\bibliography{reference}

\section*{APPENDIX}

\subsection{Proof of~\cref{prop3.1}}
\begin{proof}
For the upper bound of the LSE operator, we have
\begin{align*}
    & h_{LSE}^{\beta}(Q(s,\cdot)) \\
    =& \frac{1}{\beta} \ln \left(\sum_{a \in \mathcal{A}} \exp{\left(Q(s,a)\beta\right)}\right) \\
    \le& \frac{1}{\beta} \ln \left(\sum_{a \in \mathcal{A}} \exp{\left(Q(s,\mathrm{argmax}_{a \in \mathcal{A}}Q(s,a))\beta\right)}\right) \\
    =& \frac{1}{\beta} \ln \left(|\mathcal{A}| \exp{\left(Q(s,\mathrm{argmax}_{a \in \mathcal{A}}Q(s,a))\beta\right)}\right).
\end{align*}

The above result can be divided into two terms for all state-action pairs $(s,a)$ as
\begin{align*}
    & \frac{1}{\beta} \ln \left(|\mathcal{A}| \exp{\left(Q(s,\mathrm{argmax}_{a \in \mathcal{A}}Q(s,a))\beta\right)}\right) \\
    =& \frac{1}{\beta} \ln \left(\exp{\left(Q(s,\mathrm{argmax}_{a \in \mathcal{A}}Q(s,a))\beta\right)}\right) + \frac{\ln(|\mathcal{A}|)}{\beta} \\
    =& \max_{a \in \mathcal{A}}Q(s,a) + \frac{\ln(|\mathcal{A}|)}{\beta}.
\end{align*}

Furthermore, the lower bound of the LSE operator can be obtained as follows:
\begin{align*}
    h_{LSE}^{\beta}(Q(s,\cdot))
    =& \frac{1}{\beta} \ln \left(\sum_{a \in \mathcal{A}} \exp{\left(Q(s,a)\beta\right)}\right) \\
    \ge& \frac{1}{\beta} \ln \left( \max_{a \in \mathcal{A}} \left( \exp (Q(s,a)\beta)\right)\right) \\
    =& \frac{1}{\beta} \ln \left( \exp \left( \max_{a \in \mathcal{A}} \left( Q(s,a)\beta \right)\right)\right) \\
    =& \max_{a \in \mathcal{A}}Q(s,a).
\end{align*}
\end{proof} \vspace{0.3cm}

\subsection{Proof of~\cref{lem w lse expectation}}
\begin{proof}
Taking the expectation on~\eqref{eq: lse noise}, we have
\begin{align*}
    & \mathbb{E}\left[w_k^{LSE}\right] \\
    =& \mathbb{E}\Big[(e_{a_k} \otimes e_{s_k})r_k + \gamma(e_{a_k} \otimes e_{s_k})(e_{s_k'})^T H_{LSE}^{\beta}\left(Q_k^{LSE}\right) \\
    &- (e_{a_k} \otimes e_{s_k})(e_{a_k} \otimes e_{s_k})^T Q_k^{LSE} \\
    &- \left\{DR+\gamma DPH_{LSE}^{\beta}\left(Q_k^{LSE}\right)-DQ_k^{LSE}\right\}\Big] \\
    =& DR+\gamma DPH_{LSE}^{\beta}\left(Q_k^{LSE}\right)-DQ_k^{LSE} \\
    &- \left\{DR+\gamma DPH_{LSE}^{\beta}\left(Q_k^{LSE}\right)-DQ_k^{LSE}\right\} \\
    =& 0.
\end{align*}
\end{proof} \vspace{0.3cm}

\subsection{Proof of~\cref{lem3.12}}
Before proving~\cref{lem3.12}, we first provide some useful lemmas on the boundedness of the LSE soft Q-learning iterates (~\cref{lem2.1}), followed by the boundedness of the maximum eigenvalue of $W_k$ (~\cref{lem3.1}).
\vspace{0.3cm}
\begin{lemma} \label{lem2.1}
For all $k \ge 0$,
\[\left\|Q_k^{LSE}\right\|_{\infty} \le \frac{1}{1-\gamma}\left(1 + \gamma  \frac{\ln(|\mathcal{A}|)}{\beta}\right).\]
\end{lemma} \vspace{0.3cm}

\begin{proof}
The Q-function update equation of LSE soft Q-learning can be written as
\begin{equation}\label{eq: lem2.1 lse soft q update}
\begin{aligned}
    Q_{i+1}^{LSE} =& Q_i^{LSE} + \alpha \big\{(e_{a_i} \otimes e_{s_i})r_i \\
    &+ \gamma (e_{a_i} \otimes e_{s_i})(e_{s_i'})^T H_{LSE}^{\beta}\left(Q_i^{LSE}\right) \\
    &- (e_{a_i} \otimes e_{s_i})(e_{a_i} \otimes e_{s_i})^TQ_i^{LSE}\big\} .
\end{aligned}
\end{equation}

On both sides of~\eqref{eq: lem2.1 lse soft q update}, taking the norm with $i=0$ leads to
\begin{align*}
    & \left\|Q_{1}^{LSE}\right\|_{\infty} \\
    \le& (1-\alpha)\left\|Q_0^{LSE}\right\|_{\infty} + \alpha \left\{ \|r_0\|_{\infty} +\gamma \left\| H_{LSE}^{\beta}\left(Q_0^{LSE}\right) \right\|_{\infty} \right\} \\
    =& (1-\alpha)\left\|Q_0^{LSE}\right\|_{\infty} + \alpha \Big\{\|r_0\|_{\infty} \\
    &+ \gamma \left\| H_{LSE}^{\beta} \left(Q_0^{LSE}\right) - \Pi_{Q_0^{LSE}}^{\max} Q_0^{LSE} + \Pi_{Q_0^{LSE}}^{\max} Q_0^{LSE} \right\|_{\infty} \Big\} \\
    \le& (1-\alpha)\left\|Q_0^{LSE}\right\|_{\infty} \\
    &+ \alpha \Big\{\|r_0\|_{\infty} + \gamma \left\| H_{LSE}^{\beta} \left(Q_0^{LSE}\right) - \Pi_{Q_0^{LSE}}^{\max} Q_0^{LSE}\right\|_{\infty} \\
    &+ \gamma \left\| \Pi_{Q_0^{LSE}}^{\max} Q_0^{LSE} \right\|_{\infty} \Big\} \\
    \le& (1-\alpha)\left\|Q_0^{LSE}\right\|_{\infty} \\
    &+ \alpha \left\{\|r_0\|_{\infty} + \gamma \left\| \frac{\ln(|\mathcal{A}|)}{\beta} \textbf{1} \right\|_{\infty} + \gamma \left\| \Pi_{Q_0^{LSE}}^{\max} Q_0^{LSE} \right\|_{\infty} \right\} \\  
    \le& (1-\alpha) + \alpha + \alpha \gamma \frac{\ln(|\mathcal{A}|)}{\beta} + \alpha \gamma \\
    =& 1 + \alpha\gamma + \alpha\gamma \frac{\ln(|\mathcal{A}|)}{\beta} \\
    \le& (1 + \gamma) + \gamma\frac{\ln(|\mathcal{A}|)}{\beta},
\end{align*}
where~\cref{prop3.1} is used in the third inequality, and~\cref{as: 1} yields the fourth and last inequalities.

For applying the induction argument, suppose that
\begin{equation}\label{lse induction}
\begin{aligned}
    & \left\|Q_k^{LSE}\right\|_{\infty} \\
    \le& (1 + \gamma + \cdots + \gamma^k) + (\gamma + \cdots + \gamma^k)\frac{\ln(|\mathcal{A}|)}{\beta}
\end{aligned}
\end{equation}
holds for some $i=k-1 \ge 0$.

Then, taking the norm on~\eqref{eq: lem2.1 lse soft q update} with $i=k$ gives
\begin{align*}
    & \left\|Q_{k+1}^{LSE}\right\|_{\infty} \\
    \le& (1-\alpha)\left\|Q_k^{LSE}\right\|_{\infty} + \alpha \left\{ \|r_k\|_{\infty} +\gamma \left\| H_{LSE}^{\beta} \left(Q_k^{LSE}\right) \right\|_{\infty} \right\} \\
    =& (1-\alpha)\left\|Q_k^{LSE}\right\|_{\infty} + \alpha \Big\{ \|r_k\|_{\infty} \\
    &+ \gamma \left\| H_{LSE}^{\beta} \left(Q_k^{LSE}\right) - \Pi _{Q_k^{LSE}}^{\max}Q_k^{LSE} + \Pi _{Q_k^{LSE}}^{\max}Q_k^{LSE} \right\|_{\infty} \Big\} \\
    \le& (1-\alpha)\left\|Q_k^{LSE}\right\|_{\infty} \\
    &+ \alpha \Big\{ \|r_k\|_{\infty} + \gamma \left\| H_{LSE}^{\beta} \left(Q_k^{LSE}\right) - \Pi _{Q_k^{LSE}}^{\max}Q_k^{LSE}\right\|_{\infty} \\
    &+ \gamma \left\|\Pi _{Q_k^{LSE}}^{\max}Q_k^{LSE} \right\|_{\infty} \Big\} \\
    \le& (1-\alpha)\left\|Q_k^{LSE}\right\|_{\infty} \\
    &+ \alpha \left\{ \|r_k\|_{\infty} + \gamma \left\| \frac{\ln(|\mathcal{A}|)}{\beta} \textbf{1} \right\|_{\infty} + \gamma \left\|\Pi _{Q_k^{LSE}}^{\max}Q_k^{LSE} \right\|_{\infty} \right\} \\
    \le& (1-\alpha) \left\{(1 + \gamma + \cdots + \gamma^k) + (\gamma + \cdots + \gamma^k)\frac{\ln(|\mathcal{A}|)}{\beta} \right\} \\
    &+ \alpha\gamma \left\{(1 + \gamma + \cdots + \gamma^k) + (\gamma + \cdots + \gamma^k)\frac{\ln(|\mathcal{A}|)}{\beta} \right\} \\
    &+ \alpha + \alpha\gamma\frac{\ln(|\mathcal{A}|)}{\beta} \\
    =& (1-\alpha) \left\{(1 + \gamma + \cdots + \gamma^k) + (\gamma + \cdots + \gamma^k)\frac{\ln(|\mathcal{A}|)}{\beta} \right\} \\
    &+ \alpha \left\{(1 + \gamma + \cdots + \gamma^k) + (\gamma + \cdots + \gamma^k)\frac{\ln(|\mathcal{A}|)}{\beta} \right\} \\
    &+ \alpha \left(\gamma^{k+1} + \gamma^{k+1}\frac{\ln(|\mathcal{A}|)}{\beta} \right) \\
    =& \left\{(1 + \gamma + \cdots + \gamma^k) + (\gamma  + \cdots + \gamma^k)\frac{\ln(|\mathcal{A}|)}{\beta} \right\} \\
    &+ \alpha \left(\gamma^{k+1} + \gamma^{k+1}\frac{\ln(|\mathcal{A}|)}{\beta} \right) \\
    \le& (1 + \gamma + \cdots + \gamma^k + \gamma^{k+1}) \\
    &+ (\gamma + \cdots + \gamma^k + \gamma^{k+1})\frac{\ln(|\mathcal{A}|)}{\beta}
\end{align*}
using~\cref{prop3.1} in the third inequality, and~\cref{as: 1} and~\eqref{lse induction} in the fourth and last inequalities.

From this result, we have
\begin{align*}
    & \left\|Q_k^{LSE}\right\|_{\infty} \\
    \le& (1 + \gamma + \cdots + \gamma^k) + (\gamma + \cdots + \gamma^k)\frac{\ln(|\mathcal{A}|)}{\beta} \\
    =& (1 + \gamma + \cdots + \gamma^k) + \gamma (1 + \gamma + \cdots + \gamma^{k-1})\frac{\ln(|\mathcal{A}|)}{\beta} \\
    \le& \sum_{i=0}^{\infty} \gamma^i + \gamma \frac{\ln(|\mathcal{A}|)}{\beta}\sum_{i=0}^{\infty} \gamma^i \\
    \le& \frac{1}{1-\gamma} \left(1 + \gamma \frac{\ln(|\mathcal{A}|)}{\beta} \right),
\end{align*}
which brings the proof to a conclusion.
\end{proof} \vspace{0.3cm}

\vspace{0.3cm}
\begin{lemma}\label{lem3.1}
Using the definition of $W_k$ in~\eqref{eq: W_k}, the maximum eigenvalue of $W_k$ is bounded as
\[\lambda_{\max}(W_k) \le \frac{6\left(\ln (|\mathcal{A}|)+\beta\right)^2}{\beta^2(1-\gamma)^2}\]
for all $k \ge 0$.
\end{lemma} \vspace{0.3cm}

\begin{proof}
We have
\begin{align*}
    & \lambda_{\max}(W_k) \\
    \le& \text{tr}(W_k) \\
    =& \text{tr}\left(\mathbb{E}\left[\left(w_k^{LSE}\right) \left(w_k^{LSE}\right)^T\right]\right) \\
    =& \mathbb{E}\left[\text{tr}\left(\left(w_k^{LSE}\right) \left(w_k^{LSE}\right)^T\right)\right] \\
    =& \mathbb{E}\left[\left(w_k^{LSE}\right)^T \left(w_k^{LSE}\right)\right] \\
    =& \mathbb{E}\left[\left\|w_k^{LSE}\right\|_2^2\right] \\
    =& \mathbb{E} \bigg[\Big\|(e_{a_k} \otimes e_{s_k})r_k + \gamma(e_{a_k} \otimes e_{s_k})(e_{s_k'})^T H_{LSE}^{\beta} \left(Q_k^{LSE}\right) \\
    &- (e_{a_k} \otimes e_{s_k})(e_{a_k} \otimes e_{s_k})^T Q_k^{LSE} \\
    &- \left\{DR+\gamma DP H_{LSE}^{\beta} \left(Q_k^{LSE}\right)-DQ_k^{LSE}\right\}\Big\|_2^2 \bigg] \\
    =& \mathbb{E}\left[\left\| r_k + \gamma(e_{s_k'})^T H_{LSE}^{\beta} \left(Q_k^{LSE}\right) - (e_{a_k} \otimes e_{s_k})^T Q_k^{LSE}\right\|_2^2 \right] \\
    &- \left\| DR+\gamma DP H_{LSE}^{\beta} \left(Q_k^{LSE}\right)-DQ_k^{LSE} \right\|_2^2 \\
    \le& \mathbb{E}\left[\left\| r_k + \gamma(e_{s_k'})^T H_{LSE}^{\beta} \left(Q_k^{LSE}\right) - (e_{a_k} \otimes e_{s_k})^T Q_k^{LSE}\right\|_2^2 \right] \\
    \le& 3\mathbb{E}\left[\|r_k\|_2^2\right] + 3\mathbb{E} \left[\left\|\gamma (e_{s_k'})^T H_{LSE}^{\beta} \left(Q_k^{LSE}\right)\right\|_2^2 \right] \\
    &+ 3\mathbb{E} \left[ \left\|(e_{a_k} \otimes e_{s_k})^T Q_k^{LSE} \right\|_2^2 \right] \\
    =& 3\mathbb{E}\left[\|r_k\|_2^2\right] + 3\mathbb{E} \bigg[\Big\|\gamma (e_{s_k'})^T \cdot \\
    & \Big\{ H_{LSE}^{\beta} \left(Q_k^{LSE}\right) - {\Pi_{Q_k^{LSE}}^{\max}}Q_k^{LSE} + {\Pi_{Q_k^{LSE}}^{\max}}Q_k^{LSE} \Big\}\Big\|_2^2 \bigg] \\
    &+ 3\mathbb{E} \left[ \left\|(e_{a_k} \otimes e_{s_k})^T Q_k^{LSE} \right\|_2^2 \right] \\
    \le& 3\mathbb{E}\left[\|r_k\|_2^2\right] \\
    &+ 3\mathbb{E} \bigg[\Big\|\gamma (e_{s_k'})^T \Big\{ H_{LSE}^{\beta} \left(Q_k^{LSE}\right) - {\Pi_{Q_k^{LSE}}^{\max}}Q_k^{LSE} \Big\} \Big\|_2^2 \bigg] \\
    &+ 6\mathbb{E} \bigg[\Big| \gamma (e_{s_k'})^T \Big\{ H_{LSE}^{\beta} \left(Q_k^{LSE}\right) - {\Pi_{Q_k^{LSE}}^{\max}}Q_k^{LSE} \Big\} \Big| \cdot \\
    & \Big| \gamma (e_{s_k'})^T \Pi_{Q_k^{LSE}}^{\max}Q_k^{LSE} \Big| \bigg] \\
    &+ 3\mathbb{E} \bigg[\Big\|\gamma (e_{s_k'})^T \Pi_{Q_k^{LSE}}^{\max} Q_k^{LSE} \Big\|_2^2 \bigg] \\
    &+ 3\mathbb{E} \left[ \left\|(e_{a_k} \otimes e_{s_k})^T Q_k^{LSE} \right\|_2^2 \right] \\
    \le& 3\mathbb{E}\left[\|r_k\|_2^2\right] + 3\mathbb{E} \bigg[\Big\|\gamma (e_{s_k'})^T \frac{{\ln (|\mathcal{A}|)}}{\beta }{\bf{1}} \Big\|_2^2 \bigg] \\
    &+ 6\mathbb{E} \bigg[\Big| \gamma (e_{s_k'})^T \frac{{\ln (|\mathcal{A}|)}}{\beta }{\bf{1}} \Big| \Big| \gamma (e_{s_k'})^T \Pi_{Q_k^{LSE}}^{\max}Q_k^{LSE} \Big| \bigg] \\
    &+ 3\mathbb{E} \bigg[\Big\|\gamma (e_{s_k'})^T \Pi_{Q_k^{LSE}}^{\max} Q_k^{LSE} \Big\|_2^2 \bigg] \\
    &+ 3\mathbb{E} \left[ \left\|(e_{a_k} \otimes e_{s_k})^T Q_k^{LSE} \right\|_2^2 \right] \\
    \le& 3 + 3\gamma^2 \left(\frac{{\ln (|\mathcal{A}|)}}{\beta }\right)^2 \\
    &+ 6\gamma^2 \cdot \frac{{\ln (|\mathcal{A}|)}}{\beta } \cdot \frac{1}{(1-\gamma)} \left(1+ \gamma \frac{{\ln (|\mathcal{A}|)}}{\beta}\right) \\
    &+ 3\gamma^2 \left\{\frac{1}{1-\gamma} \left(1+ \gamma \frac{{\ln (|\mathcal{A}|)}}{\beta}\right) \right\}^2 \\
    &+ 3 \left\{\frac{1}{1-\gamma} \left(1+ \gamma \frac{{\ln (|\mathcal{A}|)}}{\beta}\right) \right\}^2 \\
    \le& \frac{6(\ln (|\mathcal{A}|)+\beta)^2}{\beta^2(1-\gamma)^2},
\end{align*}
where the third inequality comes from $\|a+b+c\|_2^2 \le 3\|a\|_2^2 + 3\|b\|_2^2 + 3\|c\|_2^2$ with arbitrary $a, b, c$, the fifth inequality is due to~\cref{prop3.1}, and the sixth inequality derives from~\cref{as: 1,lem2.1}.
\end{proof} \vspace{0.3cm}

Now, we prove~\cref{lem3.12} using the previous lemmas.

\vspace{0.3cm}
\textbf{\textit{Lemma 4.4: }}For all $k \ge 0$, the trace of $X_k$ is bounded as
\begin{align*}
    \text{tr}(X_k) \le&  \frac{6\alpha \left(\ln (|\mathcal{A}|)+\beta\right)^2|\mathcal{S} \times \mathcal{A}|^2}{\beta^2 d_{min}(1-\gamma)^3} \\
     &+ |\mathcal{S} \times \mathcal{A}|^2 \left\|Q_0^L - {Q^*}\right\|_2^2 \rho^{2k}.
\end{align*}

\vspace{0.3cm}
\begin{proof}
Given $X_k \succeq 0$, which denotes that the diagonal elements are non-negative, we get
\begin{align*}
    & \text{tr}(X_k) \\
    \le& |\mathcal{S} \times \mathcal{A}| \lambda_{max}(X_k) \\
    \le& \alpha^2 |\mathcal{S} \times \mathcal{A}| \sum_{i=0}^{k-1} \lambda_{max} \left(A_{Q^*}^i W_{k-i-1} \left(A_{Q^*}^T \right)^i\right) \\
    &+ |\mathcal{S} \times \mathcal{A}| \lambda_{max} \left(A_{Q^*}^k X_0 \left(A_{Q^*}^T\right)^k \right)\\
    \le& \alpha^2 |\mathcal{S} \times \mathcal{A}| \sup_{j \ge 0}\lambda_{max}(W_j) \sum_{i=0}^{k-1} \lambda_{max} \left(A_{Q^*}^i \left(A_{Q^*}^T \right)^i \right) \\
    &+ |\mathcal{S} \times \mathcal{A}| \lambda_{max}(X_0) \lambda_{max} \left(A_{Q^*}^k \left(A_{Q^*}^T \right)^k \right) \\
    =& \frac{6\alpha^2 \left(\ln (|\mathcal{A}|)+\beta \right)^2 |\mathcal{S} \times \mathcal{A}|}{\beta^2(1-\gamma)^2}  \sum_{i=0}^{k-1} \left\|A_{Q^*}^i\right\|_2^2 \\
    &+ |\mathcal{S} \times \mathcal{A}| \lambda_{max}(X_0) \left\|A_{Q^*}^k\right\|_2^2 \\
    \le& \frac{6\alpha^2 \left(\ln (|\mathcal{A}|)+\beta \right)^2 |\mathcal{S} \times \mathcal{A}|^2}{\beta^2(1-\gamma)^2}  \sum_{i=0}^{k-1} \left\|A_{Q^*}^i\right\|_{\infty}^2 \\
    &+ |\mathcal{S} \times \mathcal{A}|^2 \lambda_{max}(X_0) \left\|A_{Q^*}^k\right\|_{\infty}^2 \\
    \le& \frac{6\alpha^2 \left(\ln (|\mathcal{A}|)+\beta \right)^2 |\mathcal{S} \times \mathcal{A}|^2}{\beta^2(1-\gamma)^2} \sum_{i=0}^{\infty} \rho^{2i} \\
    &+ |\mathcal{S} \times \mathcal{A}|^2 \lambda_{max}(X_0) \rho^{2k} \\
    \le& \frac{6\alpha^2 \left(\ln (|\mathcal{A}|)+\beta \right)^2 |\mathcal{S} \times \mathcal{A}|^2}{\beta^2(1-\gamma)^2} \frac{1}{1-\rho^2} \\
    &+ |\mathcal{S} \times \mathcal{A}|^2 \lambda_{max}(X_0) \rho^{2k} \\
    \le& \frac{6\alpha^2 \left(\ln (|\mathcal{A}|)+\beta \right)^2 |\mathcal{S} \times \mathcal{A}|^2}{\beta^2(1-\gamma)^2} \frac{1}{1-\rho} \\
    &+ |\mathcal{S} \times \mathcal{A}|^2 \lambda_{max}(X_0) \rho^{2k},
\end{align*}
where the second inequality is obtained by applying~\eqref{autocorrelation matrix} recursively and utilizing $A_{Q^*}^i W_{k-i-1} (A_{Q^*}^T)^i \succeq 0$, $A_{Q^*}^k X_0 (A_{Q^*}^T)^k \succeq 0$. Furthermore, the first equality derives from~\cref{lem3.1} and the fifth inequality comes from~\cref{lem3.2}.
The sixth and last inequalities is due to $\rho \in (0,1)$.

By using~\cref{def3} and the relation
\begin{align*}
    \lambda_{\max}(X_0) \le& \text{tr}(X_0) \\
    =& \text{tr} \left(\mathbb{E} \left[\left(Q_0^L - {Q^*}\right) \left(Q_0^L - {Q^*}\right)^T \right] \right) \\
    =& \text{tr} \left( \left(Q_0^L - {Q^*}\right) \left(Q_0^L - {Q^*}\right)^T \right) \\
    =& \left\|Q_0^L - {Q^*}\right\|_2^2,
\end{align*}
the desired conclusion can be obtained.
\end{proof} \vspace{0.3cm}

\subsection{Proof of~\cref{prop4}}
\begin{proof}
the upper bound of the Boltzmann operator can be obtained as follows:
\begin{align}
    h_{Boltz}^{\beta} (Q(s,\cdot))
    =& \frac{\sum_{a \in \mathcal{A}} Q(s,a) \exp(Q(s,a)\beta)}{\sum_{u \in \mathcal{A}} \exp(Q(s,u)\beta )} \nonumber \\
    \le& \frac{\max_{a \in \mathcal{A}} Q(s,a) \sum_{a \in \mathcal{A}} \exp(Q(s,a)\beta)}{\sum_{u \in \mathcal{A}} \exp(Q(s,u)\beta )} \nonumber \\
    =& \max_{a \in \mathcal{A}}Q(s,a). \label{h_bz upper}
\end{align}

In addition, for the lower bound of the Boltzmann operator, we utilize~\eqref{h_bz upper} and~\cref{prop3.1}:
\begin{align*}
    h_{Boltz}^{\beta} (Q(s,\cdot))
    =& \frac{\sum_{a \in \mathcal{A}} Q(s,a) \exp(Q(s,a)\beta)}{\sum_{u \in \mathcal{A}} \exp(Q(s,u)\beta )} \\
    \le& \max_{a \in \mathcal{A}}Q(s,a) \\
    \le& \frac{1}{\beta} \ln \left(\sum_{a \in \mathcal{A}} \exp{\left(Q(s,a)\beta\right)}\right) \\
    =& h_{LSE}^{\beta} (Q(s,\cdot)).
\end{align*}
Since $h_{LSE}^{\beta} (x)$ is a convex function in $x$, it satisfies
\[h_{LSE}^{\beta} (y) \ge h_{LSE}^{\beta} (x) + \nabla_x h_{LSE}^{\beta} (x)^T(y-x) \text{, } \forall x,y \in \mathbb{R}.\]
If we choose $x=Q(s,\cdot)$ and $y=0$, then
\begin{align*}
h_{LSE}^{\beta} (0)
=& \frac{\ln(|\mathcal{A}|)}{\beta} \\
\ge& h_{LSE}^{\beta} (Q(s,\cdot)) - \nabla_{Q(s,\cdot)} h_{LSE}^{\beta} (Q(s,\cdot))^TQ(s,\cdot) \\
\ge& \max_{a \in \mathcal{A}}Q(s,a) - \nabla_{Q(s,\cdot)} h_{LSE}^{\beta} (Q(s,\cdot))^TQ(s,\cdot) \\
=& \max_{a \in \mathcal{A}}Q(s,a) - h_{Boltz}^{\beta} (Q(s,\cdot)),
\end{align*}
where the last equality uses the fact that $\nabla_x h_{LSE}^{\beta} (x)^Tx = h_{Boltz}^{\beta} (x)$.
Therefore,
\[ h_{Boltz}^{\beta} (Q(s,\cdot)) \ge \max_{a \in \mathcal{A}}Q(s,a) - \frac{\ln(|\mathcal{A}|)}{\beta}. \]
\end{proof}

\subsection{Proof of~\cref{prop lower boltz}}
\begin{proof}
To apply the induction argument, assume that $Q_k^L - {Q^*} \le Q_k^{Boltz} - {Q^*}$ holds for some $k \ge 0$. Then, we have
\begin{align*}
    & Q_{k + 1}^{Boltz} - {Q^*} \\
    =& Q_k^{Boltz} - {Q^*} \\
    &+ \alpha \left\{ DR + \gamma DP H_{Boltz}^{\beta} \left(Q_k^{Boltz}\right) - DQ_k^{Boltz} + w_k^{Boltz} \right\} \\
    \ge& Q_k^{Boltz} - {Q^*} + \alpha DR \\
    &+ \alpha\gamma DP \left({\Pi _{Q_k^{Boltz}}^{\max}}Q_k^{Boltz} - \frac{\ln(|\mathcal{A}|)}{\beta} \textbf{1} \right) \\
    &+ \alpha \left(- DQ_k^{Boltz} + w_k^{Boltz} \right) \\
    \ge& Q_k^{Boltz} - {Q^*} + \alpha DR \\
    &+ \alpha\gamma DP \left({\Pi _{Q^*}^{\max}}Q_k^{Boltz} - \frac{\ln(|\mathcal{A}|)}{\beta} \textbf{1} \right) \\
    &+ \alpha \left(- DQ_k^{Boltz} + w_k^{Boltz} \right) \\
    =& Q_k^{Boltz} - {Q^*} - \alpha \left(\gamma DP {\Pi _{Q^*}^{\max}}-D \right)Q^* \\
    &+ \alpha\gamma DP \left({\Pi _{Q^*}^{\max}}Q_k^{Boltz} - \frac{\ln(|\mathcal{A}|)}{\beta} \textbf{1} \right) \\
    &+ \alpha \left(- DQ_k^{Boltz} + w_k^{Boltz} \right) \\
    =& \left\{I+ \alpha \left(\gamma DP{\Pi _{Q^*}^{\max}} - D\right)\right\}\left(Q_k^{Boltz} - {Q^*}\right) + \alpha w_k^{Boltz} \\
    &- \alpha\gamma DP \frac{\ln(|\mathcal{A}|)}{\beta} \textbf{1} \\
    \ge& \left\{I+ \alpha \left(\gamma DP{\Pi _{Q^*}^{\max}} - D \right)\right\}\left(Q_k^L - {Q^*}\right) + \alpha w_k^{Boltz} \\
    &- \alpha\gamma DP \frac{\ln(|\mathcal{A}|)}{\beta} \textbf{1} \\
    =& Q_{k+1}^L - {Q^*},
\end{align*}
where~\cref{prop4} is used in the first inequality and the optimal Bellman equation is utilized in the second equality. The last inequality relies on the assumption $Q_k^L - {Q^*} \le Q_k^{Boltz} - {Q^*}$ and the fact that all elements of $I+ \alpha \left(\gamma DP{\Pi _{Q^*}^{\max}} - D\right)$ are non-negative. This completes the proof.
\end{proof} \vspace{0.3cm}

\subsection{Proof of~\cref{thm3.3}}
In order to prove~\cref{thm3.3}, we first present three helpful lemmas on the boundedness of the Boltzmann soft Q-learning iterates (~\cref{lem5.1}), and some features of the stochastic noise~\eqref{eq: boltz noise} (~\cref{lem3.4,lem3.44}).

\vspace{0.3cm}
\begin{lemma} \label{lem5.1}
For all $k \ge 0$,
\[\left\|Q_k^{Boltz}\right\|_{\infty} \le \frac{1}{1-\gamma}\left(1 + \gamma  \frac{\ln(|\mathcal{A}|)}{\beta}\right).\]
\end{lemma} \vspace{0.3cm}

\begin{proof}
The Q-function update equation of Boltzmann soft Q-learning can be written as
\begin{equation}\label{eq: lem2.1 boltz soft q update}
\begin{aligned}
    Q_{i+1}^{Boltz} =& Q_i^{Boltz} + \alpha \Big\{(e_{a_i} \otimes e_{s_i})r_i \\
    &+ \gamma (e_{a_i} \otimes e_{s_i})(e_{s_i'})^T H_{Boltz}^{\beta}\left( Q_i^{Boltz} \right) \\
    &- (e_{a_i} \otimes e_{s_i})(e_{a_i} \otimes e_{s_i})^T Q_i^{Boltz}\Big\}.
\end{aligned}
\end{equation}

On both sides of~\eqref{eq: lem2.1 boltz soft q update}, taking the norm with $i=0$ leads to
\begin{align*}
    & \left\|Q_{1}^{Boltz}\right\|_{\infty} \\
    \le& (1-\alpha)\left\|Q_0^{Boltz}\right\|_{\infty} \\
    &+ \alpha \left\{ \|r_0\|_{\infty} +\gamma \left\| H_{Boltz}^{\beta}\left( Q_0^{Boltz} \right) \right\|_{\infty} \right\} \\
    =& (1-\alpha)\left\|Q_0^{Boltz}\right\|_{\infty} + \alpha \Big\{ \|r_0\|_{\infty} + \gamma \Big\| H_{Boltz}^{\beta}\left( Q_0^{Boltz} \right) \\
    &- \Pi_{Q_0^{Boltz}}^{\max} Q_0^{Boltz} + \Pi_{Q_0^{Boltz}}^{\max} Q_0^{Boltz} \Big\|_{\infty} \Big\} \\
    \le& (1-\alpha)\left\|Q_0^{Boltz}\right\|_{\infty} \\
    &+ \alpha \Big\{ \|r_0\|_{\infty} + \gamma \Big\| H_{Boltz}^{\beta}\left( Q_0^{Boltz} \right) - \Pi_{Q_0^{Boltz}}^{\max} Q_0^{Boltz} \Big\|_{\infty} \\
    &+ \gamma \Big\| \Pi_{Q_0^{Boltz}}^{\max} Q_0^{Boltz} \Big\|_{\infty} \Big\} \\
    \le& (1-\alpha)\left\|Q_0^{Boltz}\right\|_{\infty} \\
    &+ \alpha \left( \|r_0\|_{\infty} +  \gamma \left\| \frac{\ln(|\mathcal{A}|)}{\beta} \textbf{1} \right\|_{\infty} + \gamma \Big\| \Pi_{Q_0^{Boltz}}^{\max} Q_0^{Boltz} \Big\|_{\infty} \right) \\
    \le& (1-\alpha) + \alpha + \alpha \gamma \frac{\ln(|\mathcal{A}|)}{\beta} + \alpha \gamma \\
    =& 1 + \alpha\gamma + \alpha\gamma \frac{\ln(|\mathcal{A}|)}{\beta} \\
    \le& (1 + \gamma) + \gamma\frac{\ln(|\mathcal{A}|)}{\beta},
\end{align*}
where~\cref{prop4} is used in the third inequality, and~\cref{as: 1} yields the fourth and last inequalities.

For applying the induction argument, suppose that
\begin{equation}\label{boltz induction}
\begin{aligned}
    & \left\|Q_k^{Boltz}\right\|_{\infty} \\
    \le& (1 + \gamma + \cdots + \gamma^k) + (\gamma + \cdots + \gamma^k)\frac{\ln(|\mathcal{A}|)}{\beta}
\end{aligned}
\end{equation}
holds for some $i=k-1 \ge 0$.

Then, taking the norm on~\eqref{eq: lem2.1 boltz soft q update} with $i=k$ gives
\begin{align*}
    & \left\|Q_{k+1}^{Boltz}\right\|_{\infty} \\
    \le& (1-\alpha)\left\|Q_k^{Boltz}\right\|_{\infty} \\
    &+ \alpha \left\{\|r_k\|_{\infty} +\gamma \left\| H_{Boltz}^{\beta}\left( Q_k^{Boltz} \right) \right\|_{\infty} \right\} \\
    =& (1-\alpha)\left\|Q_k^{Boltz}\right\|_{\infty} + \alpha \Big\{ \|r_k\|_{\infty} +\gamma \Big\| H_{Boltz}^{\beta}\left( Q_k^{Boltz} \right) \\
    &- \Pi _{Q_k^{Boltz}}^{\max}Q_k^{Boltz} + \Pi _{Q_k^{Boltz}}^{\max}Q_k^{Boltz} \Big\|_{\infty} \Big\} \\
    \le& (1-\alpha)\left\|Q_k^{Boltz}\right\|_{\infty} \\
    &+ \alpha \Big\{ \|r_k\|_{\infty} +\gamma \Big\| H_{Boltz}^{\beta}\left( Q_k^{Boltz} \right) - \Pi _{Q_k^{Boltz}}^{\max}Q_k^{Boltz} \Big\|_{\infty} \\
    &+ \gamma \Big\| \Pi _{Q_k^{Boltz}}^{\max}Q_k^{Boltz} \Big\|_{\infty} \Big\} \\
    \le& (1-\alpha)\left\|Q_k^{Boltz}\right\|_{\infty} \\
    &+ \alpha \left\{ \|r_k\|_{\infty} + \gamma \left\| \frac{\ln(|\mathcal{A}|)}{\beta} \textbf{1} \right\|_{\infty} + \gamma \Big\| \Pi_{Q_k^{Boltz}}^{\max} Q_k^{Boltz} \Big\|_{\infty} \right\} \\
    \le& (1-\alpha) \left\{(1 + \gamma + \cdots + \gamma^k) + (\gamma + \cdots + \gamma^k)\frac{\ln(|\mathcal{A}|)}{\beta} \right\} \\
    &+ \alpha\gamma \left\{(1 + \gamma + \cdots + \gamma^k) + (\gamma + \cdots + \gamma^k)\frac{\ln(|\mathcal{A}|)}{\beta} \right\} \\
    &+ \alpha + \alpha\gamma\frac{\ln(|\mathcal{A}|)}{\beta} \\
    =& (1-\alpha) \left\{(1 + \gamma + \cdots + \gamma^k) + (\gamma + \cdots + \gamma^k)\frac{\ln(|\mathcal{A}|)}{\beta} \right\} \\
    &+ \alpha \left\{(1 + \gamma + \cdots + \gamma^k) + (\gamma + \cdots + \gamma^k)\frac{\ln(|\mathcal{A}|)}{\beta} \right\} \\
    &+ \alpha \left(\gamma^{k+1} + \gamma^{k+1}\frac{\ln(|\mathcal{A}|)}{\beta} \right) \\
    =& \left\{(1 + \gamma + \cdots + \gamma^k) + (\gamma  + \cdots + \gamma^k)\frac{\ln(|\mathcal{A}|)}{\beta} \right\} \\
    &+ \alpha \left(\gamma^{k+1} + \gamma^{k+1}\frac{\ln(|\mathcal{A}|)}{\beta} \right) \\
    \le& (1 + \gamma + \cdots + \gamma^k + \gamma^{k+1}) \\
    &+ (\gamma + \cdots + \gamma^k + \gamma^{k+1})\frac{\ln(|\mathcal{A}|)}{\beta}
\end{align*}
using~\cref{prop4} in the third inequality, and~\cref{as: 1} and~\eqref{boltz induction} in the fourth and last inequalities.

From this result, we have
\begin{align*}
    & \left\|Q_k^{Boltz}\right\|_{\infty} \\
    \le& (1 + \gamma + \cdots + \gamma^k) + (\gamma + \cdots + \gamma^k)\frac{\ln(|\mathcal{A}|)}{\beta} \\
    =& (1 + \gamma + \cdots + \gamma^k) + \gamma (1 + \gamma + \cdots + \gamma^{k-1})\frac{\ln(|\mathcal{A}|)}{\beta} \\
    \le& \sum_{i=0}^{\infty} \gamma^i + \gamma \frac{\ln(|\mathcal{A}|)}{\beta}\sum_{i=0}^{\infty} \gamma^i \\
    \le& \frac{1}{1-\gamma} \left(1 + \gamma \frac{\ln(|\mathcal{A}|)}{\beta} \right),
\end{align*}
which completes the proof.
\end{proof} \vspace{0.3cm}

\vspace{0.3cm}
\begin{lemma} \label{lem3.4}
For any $k \ge 0$, we have
\[\mathbb{E} \left[w_k^{Boltz}\right]=0.\]
\end{lemma} \vspace{0.3cm}

\begin{proof}
Taking the expectation on~\eqref{eq: boltz noise} leads to
\begin{align*}
    & \mathbb{E} \left[w_k^{Boltz}\right] \\
    =& \mathbb{E} \Big[(e_{a_k} \otimes e_{s_k})r_k + \gamma(e_{a_k} \otimes e_{s_k})(e_{s_k'})^T H_{Boltz}^{\beta} \left( Q_k^{Boltz} \right) \\
    &- (e_{a_k} \otimes e_{s_k})(e_{a_k} \otimes e_{s_k})^T Q_k^{Boltz} \\
    &- \left\{ DR+\gamma DP H_{Boltz}^{\beta} \left( Q_k^{Boltz} \right) -DQ_k^{Boltz} \right\} \Big] \\
    =& DR+\gamma DP H_{Boltz}^{\beta} \left( Q_k^{Boltz} \right) -DQ_k^{Boltz} \\
    &- \left\{ DR+\gamma DP H_{Boltz}^{\beta} \left( Q_k^{Boltz} \right) -DQ_k^{Boltz} \right\} \\
    =& 0.
\end{align*}
\end{proof}

\vspace{0.3cm}
\begin{lemma} \label{lem3.44}
For any $k \ge 0$, we have
\[ \mathbb{E}\left[\left(w_k^{Boltz}\right)^T\left(w_k^{Boltz}\right)\right] \le  \frac{6(\ln (|\mathcal{A}|)+\beta)^2}{\beta^2(1-\gamma)^2}.\]
\end{lemma} \vspace{0.3cm}

\begin{proof}
We have
\begin{align*}
    & \mathbb{E} \left[ \left(w_k^{Boltz}\right)^T \left(w_k^{Boltz}\right) \right] \\
    =& \mathbb{E} \left[ \left\|w_k^{Boltz}\right\|_2^2 \right] \\
    =& \mathbb{E}\bigg[ \Big\|(e_{a_k} \otimes e_{s_k})r_k \\
    &+ \gamma(e_{a_k} \otimes e_{s_k})(e_{s_k'})^T H_{Boltz}^{\beta} \left( Q_k^{Boltz} \right) \\
    &- (e_{a_k} \otimes e_{s_k})(e_{a_k} \otimes e_{s_k})^T Q_k^{Boltz} \\
    &- \left\{ DR+\gamma DP H_{Boltz}^{\beta} \left( Q_k^{Boltz} \right) -DQ_k^{Boltz} \right\} \Big\|_2^2 \bigg] \\
    =& \mathbb{E} \bigg[ \Big\| r_k + \gamma(e_{s_k'})^T H_{Boltz}^{\beta} \left( Q_k^{Boltz} \right) \\
    &- (e_{a_k} \otimes e_{s_k})^T Q_k^{Boltz} \Big\|_2^2 \bigg] \\
    &- \left\|DR+\gamma DP H_{Boltz}^{\beta} \left( Q_k^{Boltz} \right) -DQ_k^{Boltz} \right\|_2^2 \\
    \le& \mathbb{E} \bigg[\Big\|r_k + \gamma(e_{s_k'})^T H_{Boltz}^{\beta} \left( Q_k^{Boltz} \right) \\
    &- (e_{a_k} \otimes e_{s_k})^T Q_k^{Boltz} \Big\|_2^2 \bigg] \\
    \le& 3\mathbb{E} \left[ \|r_k\|_2^2 \right] + 3\mathbb{E} \left[ \left\|\gamma (e_{s_k'})^T H_{Boltz}^{\beta} \left( Q_k^{Boltz}\right) \right\|_2^2 \right] \\
    &+ 3\mathbb{E} \left[ \left\|(e_{a_k} \otimes e_{s_k})^T Q_k^{Boltz} \right\|_2^2 \right] \\ 
    =& 3\mathbb{E}\left[\|r_k\|_2^2\right] + 3\mathbb{E} \bigg[\Big\|\gamma (e_{s_k'})^T \Big\{ H_{Boltz}^{\beta} \left(Q_k^{Boltz}\right) \\
    &- {\Pi_{Q_k^{Boltz}}^{\max}}Q_k^{Boltz} + {\Pi_{Q_k^{Boltz}}^{\max}}Q_k^{Boltz} \Big\}\Big\|_2^2 \bigg] \\
    &+ 3\mathbb{E} \left[ \left\|(e_{a_k} \otimes e_{s_k})^T Q_k^{Boltz} \right\|_2^2 \right] \\
    \le& 3\mathbb{E}\left[\|r_k\|_2^2\right] \\
    &+ 3\mathbb{E} \bigg[\Big\|\gamma (e_{s_k'})^T \Big\{ H_{Boltz}^{\beta} \left(Q_k^{Boltz}\right) - {\Pi_{Q_k^{Boltz}}^{\max}}Q_k^{Boltz} \Big\} \Big\|_2^2 \bigg] \\
    &+ 6\mathbb{E} \bigg[\Big| \gamma (e_{s_k'})^T \Big\{ H_{Boltz}^{\beta} \left(Q_k^{Boltz}\right) - {\Pi_{Q_k^{Boltz}}^{\max}}Q_k^{Boltz} \Big\} \Big| \cdot \\
    & \Big| \gamma (e_{s_k'})^T \Pi_{Q_k^{Boltz}}^{\max}Q_k^{Boltz} \Big| \bigg] \\
    &+ 3\mathbb{E} \bigg[\Big\|\gamma (e_{s_k'})^T \Pi_{Q_k^{Boltz}}^{\max} Q_k^{Boltz} \Big\|_2^2 \bigg] \\
    &+ 3\mathbb{E} \left[ \left\|(e_{a_k} \otimes e_{s_k})^T Q_k^{Boltz} \right\|_2^2 \right] \\
    \le& 3\mathbb{E}\left[\|r_k\|_2^2\right] + 3\mathbb{E} \bigg[\Big\|\gamma (e_{s_k'})^T \frac{{\ln (|\mathcal{A}|)}}{\beta }{\bf{1}} \Big\|_2^2 \bigg] \\
    &+ 6\mathbb{E} \bigg[\Big| \gamma (e_{s_k'})^T \frac{{\ln (|\mathcal{A}|)}}{\beta }{\bf{1}} \Big| \Big| \gamma (e_{s_k'})^T \Pi_{Q_k^{Boltz}}^{\max}Q_k^{Boltz} \Big| \bigg] \\
    &+ 3\mathbb{E} \bigg[\Big\|\gamma (e_{s_k'})^T \Pi_{Q_k^{Boltz}}^{\max} Q_k^{Boltz} \Big\|_2^2 \bigg] \\
    &+ 3\mathbb{E} \left[ \left\|(e_{a_k} \otimes e_{s_k})^T Q_k^{Boltz} \right\|_2^2 \right] \\
    \le& 3 + 3\gamma^2 \left(\frac{{\ln (|\mathcal{A}|)}}{\beta }\right)^2 \\
    &+ 6\gamma^2 \cdot \frac{{\ln (|\mathcal{A}|)}}{\beta } \cdot \frac{1}{(1-\gamma)} \left(1+ \gamma \frac{{\ln (|\mathcal{A}|)}}{\beta}\right) \\
    &+ 3\gamma^2 \left\{\frac{1}{1-\gamma} \left(1+ \gamma \frac{{\ln (|\mathcal{A}|)}}{\beta}\right) \right\}^2 \\
    &+ 3 \left\{\frac{1}{1-\gamma} \left(1+ \gamma \frac{{\ln (|\mathcal{A}|)}}{\beta}\right) \right\}^2 \\
    \le& \frac{6(\ln (|\mathcal{A}|)+\beta)^2}{\beta^2(1-\gamma)^2},
\end{align*}
where the second inequality comes from $\|a+b+c\|_2^2 \le 3\|a\|_2^2 + 3\|b\|_2^2 + 3\|c\|_2^2$ with arbitrary $a, b, c$, the fourth inequality is due to~\cref{prop4}, and the fifth inequality derives from~\cref{as: 1,lem5.1}.
\end{proof} \vspace{0.3cm}

We can now prove~\cref{thm3.3}.

\vspace{0.3cm}
\textbf{\textit{Theorem 5.1: }}For any $k \ge 0$, we have
\begin{align*}
    \mathbb{E}\left[\left\|Q_k^L - Q^*\right\|_2\right] &\le |\mathcal{S} \times \mathcal{A}|^\frac{1}{2} \left\|Q_0^L-Q^*\right\|_2 \rho^{k} \\
    &+ \frac{\sqrt{6} \alpha^\frac{1}{2} \left(\ln (|\mathcal{A}|)+\beta\right) |\mathcal{S} \times \mathcal{A}|^\frac{1}{2}}{\beta d_{min}^\frac{1}{2}(1-\gamma)^\frac{3}{2}} \\
    &+ \frac {\gamma d_{max} \ln(|\mathcal{A}|)|\mathcal{S} \times \mathcal{A}|^\frac{1}{2}}{\beta d_{min}(1-\gamma)}.
\end{align*}

\vspace{0.3cm}
\begin{proof}
Applying~\eqref{eq: Boltzmann soft q-learning lower comparison system} recursively, we get
\begin{align*}
    Q_k^L-Q^* =& A_{Q^*}^k\left(Q_0^L-Q^*\right) + \alpha \sum_{i=0}^{k-1} A_{Q^*}^{k-i-1}w_i^{Boltz} \\
    &- \alpha \gamma \sum_{i=0}^{k-1} A_{Q^*}^{k-i-1} DP \frac{\ln(|\mathcal{A}|)}{\beta}\textbf{1}.
\end{align*}

Then, taking the norm and expectation of the above equality gives
\begin{align*}
    & \mathbb{E}\left[\left\| Q_k^L-Q^*\right\|_2\right] \\
    =& \mathbb{E}\Bigg[\Bigg\| A_{Q^*}^k \left(Q_0^L-Q^*\right) + \alpha \sum_{i=0}^{k-1} A_{Q^*}^{k-i-1}w_i^{Boltz} \\
    &- \alpha \gamma \sum_{i=0}^{k-1} A_{Q^*}^{k-i-1} DP \frac{\ln(|\mathcal{A}|)}{\beta}\textbf{1}\Bigg\|_2\Bigg] \\
    \le& \mathbb{E}\left[\left\|A_{Q^*}^k\left(Q_0^L-Q^*\right)+\alpha \sum_{i=0}^{k-1} A_{Q^*}^{k-i-1}w_i^{Boltz}\right\|_2\right] \\
    &+ \mathbb{E}\left[\left\|\alpha \gamma \sum_{i=0}^{k-1} A_{Q^*}^{k-i-1} DP \frac{\ln(|\mathcal{A}|)}{\beta}\textbf{1}\right\|_2\right].
\end{align*}

By using the relation $\mathbb{E}[\|\cdot\|_2] = \mathbb{E}[\sqrt{\|\cdot\|_2^2}] \le \sqrt{\mathbb{E}[\|\cdot\|_2^2]}$, the last inequality turns into 
\begin{align*}
    & \mathbb{E}\left[\left\| Q_k^L-Q^*\right\|_2\right] \\
    \le& \sqrt{\mathbb{E}\left[\left\|A_{Q^*}^k \left(Q_0^L-Q^*\right)+\alpha \sum_{i=0}^{k-1} A_{Q^*}^{k-i-1}w_i^{Boltz}\right\|_2^2\right]} \\
    &+ \mathbb{E}\left[\left\|\alpha \gamma \sum_{i=0}^{k-1} A_{Q^*}^{k-i-1} DP \frac{\ln(|\mathcal{A}|)}{\beta}\textbf{1}\right\|_2\right].
\end{align*}

Since
\begin{align*}
    & \mathbb{E}\left[\left\|A_{Q^*}^k\left(Q_0^L-Q^*\right)+\alpha \sum_{i=0}^{k-1} A_{Q^*}^{k-i-1}w_i^{Boltz}\right\|_2^2\right] \\
    =& \mathbb{E}\Bigg[\left(A_{Q^*}^k \left(Q_0^L-Q^*\right)+\alpha \sum_{i=0}^{k-1} A_{Q^*}^{k-i-1}w_i^{Boltz} \right)^T \\
    &\cdot \left(A_{Q^*}^k \left(Q_0^L-Q^*\right)+\alpha \sum_{i=0}^{k-1} A_{Q^*}^{k-i-1}w_i^{Boltz}\right)\Bigg] \\
    =& \mathbb{E}\left[\left(Q_0^L-Q^*\right)^T\left(A_{Q^*}^k\right)^TA_{Q^*}^k\left(Q_0^L-Q^*\right)\right] \\
    &+ \mathbb{E}\left[\alpha^2 \sum_{i=0}^{k-1} \left(w_i^{Boltz}\right)^T\left(A_{Q^*}^{k-i-1}\right)^TA_{Q^*}^{k-i-1} w_i^{Boltz}\right] \\
    \le& \mathbb{E}\left[\lambda_{\max} \left( \left(A_{Q^*}^k\right)^TA_{Q^*}^k \right) \left(Q_0^L-Q^*\right)^T\left(Q_0^L-Q^*\right)\right] \\
    &+ \mathbb{E} \Bigg[ \alpha^2 \sum_{i=0}^{k-1} \lambda_{\max}\left(\left(A_{Q^*}^{k-i-1}\right)^TA_{Q^*}^{k-i-1}\right) \\
    &\cdot \left(w_i^{Boltz}\right)^T w_i^{Boltz} \Bigg] \\
    =& \left\|A_{Q^*}^k\right\|_2^2 \left\|Q_0^L-Q^*\right\|_2^2 \\
    &+ \mathbb{E}\left[ \alpha^2 \sum_{i=0}^{k-1} \left\|A_{Q^*}^{k-i-1}\right\|_2^2 \left(w_i^{Boltz}\right)^T w_i^{Boltz} \right]
\end{align*}
holds by using~\cref{lem3.4} in the second equality, we have
\begin{align*}
    & \mathbb{E}\left[\left\| Q_k^L-Q^*\right\|_2\right] \\
    \le& \Bigg( \left\|A_{Q^*}^k\right\|_2^2 \left\|Q_0^L-Q^*\right\|_2^2 \\
    &+ \mathbb{E}\left[\alpha^2 \sum_{i=0}^{k-1} \left\|A_{Q^*}^{k-i-1}\right\|_2^2 \left(w_i^{Boltz}\right)^T w_i^{Boltz}\right] \Bigg)^\frac{1}{2} \\
    &+ \mathbb{E}\left[\left\|\alpha \gamma \sum_{i=0}^{k-1} A_{Q^*}^{k-i-1} DP \frac{\ln(|\mathcal{A}|)}{\beta}\textbf{1}\right\|_2\right].
\end{align*}

Combining the previous inequality with $\|\cdot\|_2 \le |\mathcal{S} \times \mathcal{A}|^\frac{1}{2}\|\cdot\|_{\infty}$ yields
\begin{align*}
    & \mathbb{E}\left[\left\| Q_k^L-Q^*\right\|_2\right] \\
    \le& \Bigg(|\mathcal{S} \times \mathcal{A}| \left\|A_{Q^*}^k\right\|_{\infty}^2 \left\|Q_0^L-Q^*\right\|_2^2 \\
    &+ \mathbb{E}\left[\alpha^2 \sum_{i=0}^{k-1} |\mathcal{S} \times \mathcal{A}| \left\|A_{Q^*}^{k-i-1}\right\|_{\infty}^2 \left(w_i^{Boltz}\right)^T w_i^{Boltz}\right] \Bigg)^\frac{1}{2} \\
    &+ \frac {\alpha \gamma d_{max}\ln(|\mathcal{A}|) |\mathcal{S} \times \mathcal{A}|^\frac{1}{2}}{\beta}\sum_{i=0}^{k-1} \left\|A_{Q^*}^{k-i-1}\right\|_{\infty} \\
    \le& \Big(|\mathcal{S} \times \mathcal{A}| \left\|Q_0^L-Q^*\right\|_2^2 \rho^{2k} \\
    &+ \alpha^2 |\mathcal{S} \times \mathcal{A}| \sum_{i=0}^{k-1} \rho^{2(k-i-1)}\mathbb{E}\left[\left(w_i^{Boltz}\right)^T w_i^{Boltz}\right] \Big)^\frac{1}{2} \\
    &+ \frac {\alpha \gamma d_{max}\ln(|\mathcal{A}|) |\mathcal{S} \times \mathcal{A}|^\frac{1}{2}}{\beta}\sum_{i=0}^{k-1} \rho^{k-i-1},
\end{align*}
where the second inequality comes from~\cref{lem3.2}.

Noting the relations $\sum_{i=0}^{k-1} \rho^{i} \le \sum\limits_{i = 0}^\infty  {{\rho ^i}} \le \frac{1}{1 - \rho}$, $\sum_{i=0}^{k-1} \rho^{2i} \le \sum\limits_{i = 0}^\infty  {\rho ^{2i}} \le \frac{1}{1 - \rho^2} \le \frac{1}{1 - \rho}$ with~\cref{def3}, and applying~\cref{lem3.44} to the last inequality, one gets
\begin{align*}
    \mathbb{E}\left[\left\| Q_k^L-Q^*\right\|_2\right] \le& \bigg(|\mathcal{S} \times \mathcal{A}| \left\|Q_0^L-Q^*\right\|_2^2 \rho^{2k} \\
    &+ \frac{6\alpha \left(\ln (|\mathcal{A}|)+\beta\right)^2 |\mathcal{S} \times \mathcal{A}|}{\beta^2 d_{min}(1-\gamma)^3} \bigg)^\frac{1}{2} \\
    &+ \frac {\gamma d_{max} \ln(|\mathcal{A}|)|\mathcal{S} \times \mathcal{A}|^\frac{1}{2}}{\beta d_{min}(1-\gamma)}.
\end{align*}

Using the subadditivity of the square root function in the aforementioned inequality, we can obtain the final result.
\end{proof} \vspace{0.3cm}

\subsection{Proof of~\cref{prop upper boltz}}
\begin{proof}
To apply the induction argument, assume that $Q_k^U-Q^* \ge Q_k^{Boltz}-Q^*$ holds for some $k \ge 0$. Then, we have
\begin{align*}
    & Q_{k+1}^{Boltz}-Q^* \\
    =& Q_k^{Boltz} - {Q^*} \\
    &+ \alpha \left\{ DR + \gamma DP H_{Boltz}^{\beta} \left( Q_k^{Boltz} \right) - DQ_k^{Boltz} + w_k^{Boltz} \right\} \\   
    \le& Q_k^{Boltz} - {Q^*} \\
    &+ \alpha \left(DR + \gamma DP{\Pi _{Q_k^{Boltz}}^{\max}}Q_k^{boltz} - DQ_k^{Boltz} + w_k^{Boltz} \right)\\
    =& Q_k^{Boltz} - {Q^*} -\alpha \left(\gamma DP{\Pi _{Q^*}^{\max}} - D \right)Q^*\\
    &+ \alpha\gamma DP \left({\Pi _{Q_k^{Boltz}}^{\max}}Q_k^{boltz} + \Pi _{Q_k^{Boltz}}^{\max}Q^* - \Pi _{Q_k^{Boltz}}^{\max}Q^* \right) \\
    &+ \alpha \left(- DQ_k^{Boltz} + w_k^{Boltz} \right)\\
    =& \left\{I+ \alpha \left(\gamma DP{\Pi _{Q_k^{Boltz}}^{\max}} - D \right)\right\} \left(Q_k^{Boltz} - {Q^*}\right) \\
    &+ \alpha\gamma DP \left(\Pi _{Q_k^{Boltz}}^{\max} - \Pi _{Q^*}^{\max} \right)Q^* + \alpha w_k^{Boltz} \\
    \le& \left\{I+ \alpha \left(\gamma DP{\Pi _{Q_k^{Boltz}}^{\max}} - D \right)\right\} \left(Q_k^{Boltz} - {Q^*}\right) + \alpha w_k^{Boltz} \\
    \le& \left\{I+ \alpha \left(\gamma DP{\Pi _{Q_k^{Boltz}}^{\max}} - D \right)\right\} \left(Q_k^U - {Q^*}\right) + \alpha w_k^{Boltz} \\
    =& Q_{k+1}^U-Q^*,
\end{align*}
where~\cref{prop4} is used in the first inequality and the optimal Bellman equation is utilized in the second equality. The second inequality holds because $\alpha \gamma DP \left(\Pi _{Q_k^{Boltz}}^{\max}-\Pi _{Q^*}^{\max} \right)Q^* \le $$\alpha \gamma DP \left(\Pi _{Q^*}^{\max}-\Pi _{Q^*}^{\max} \right)Q^*=0$. Moreover, the last inequality relies on the assumption $Q_k^U-Q^* \ge Q_k^{Boltz}-Q^*$ and the fact that all elements of $I+ \alpha \left(\gamma DP{\Pi _{Q_k^{Boltz}}^{\max}} - D \right)$ are non-negative. This brings the proof to its conclusion.
\end{proof} \vspace{0.3cm}

\subsection{Proof of~\cref{boltzmann final upper}}
\begin{proof}
Considering the norm of~\eqref{eq: Boltzmann subtracted upper comparison system}, we obtain
\begin{align}
    & \left\|Q_{i+1}^U-Q_{i+1}^L\right\|_{\infty} \nonumber \\
    \le& \left\|A_{Q_i^{Boltz}}\right\|_{\infty} \left\|Q_i^U-Q_i^L\right\|_{\infty} + \left\|B_{Q_i^{Boltz}}\right\|_{\infty} \left\|Q_i^L-Q^*\right\|_{\infty} \nonumber \\
    &+ \left\| \alpha\gamma DP \frac{\ln(|\mathcal{A}|)}{\beta} \textbf{1} \right\|_{\infty} \nonumber \\
    \le& \rho \left\|Q_i^U-Q_i^L\right\|_{\infty} \nonumber \\
    &+ 2\alpha\gamma d_{max} \left\|Q_i^L-Q^*\right\|_{\infty} + \frac {\alpha}{\beta} \ln(|\mathcal{A}|) \label{boltz upper minus lower}
\end{align}
using~\cref{lem3.2} and the following results:
\begin{align*}
    \left\|B_{Q_i^{Boltz}}\right\|_{\infty} =& \left\|A_{Q_i^{Boltz}}-A_{Q^*}\right\|_{\infty} \\
    \le& \alpha \gamma d_{max} \left\|P\left(\Pi_{Q_i^{Boltz}}^{max}-\Pi _{Q^*}^{\max}\right)\right\|_{\infty} \\
    \le& 2\alpha \gamma d_{max}.
\end{align*}
and
\begin{align*}
    \left\| \alpha\gamma DP \frac{\ln(|\mathcal{A}|)}{\beta} \textbf{1} \right\|_{\infty} 
    \le& \alpha \gamma \|D\|_{\infty} \|P\|_{\infty} \left\|\frac {\ln(|\mathcal{A}|)}{\beta} \textbf{1}\right\|_{\infty} \\
    \le& \alpha \gamma d_{max} \frac {\ln(|\mathcal{A}|)}{\beta} \\
    \le& \frac {\alpha}{\beta} \ln(|\mathcal{A}|).
\end{align*}

Taking the expectation on~\eqref{boltz upper minus lower} and applying~\cref{thm3.3} leads to
\begin{align*}
    & \mathbb{E}\left[\left\|Q_{i+1}^U-Q_{i+1}^L\right\|_{\infty}\right] \\
    \le& \rho\mathbb{E}\left[\left\|Q_i^U-Q_i^L\right\|_{\infty}\right] + 2\alpha\gamma d_{max}\mathbb{E}\left[\left\|Q_i^L-Q^*\right\|_2\right] \\
    &+ \frac {\alpha}{\beta} \ln(|\mathcal{A}|) \\
    \le& \rho E\left[\left\|Q_i^U-Q_i^L\right\|_{\infty}\right] + 2\alpha\gamma d_{max}|\mathcal{S} \times \mathcal{A}|^\frac{1}{2}\left\|Q_0^L-Q^*\right\|_2 \rho^{i}\\
    &+ \frac{2\sqrt{6} \alpha^\frac{3}{2}\gamma d_{max} \left(\ln (|\mathcal{A}|)+\beta\right) |\mathcal{S} \times \mathcal{A}|^\frac{1}{2}}{\beta d_{min}^\frac{1}{2}(1-\gamma)^\frac{3}{2}} \\
    &+ \frac {2\alpha\gamma^2 d_{max}^2 \ln(|\mathcal{A}|) |\mathcal{S} \times \mathcal{A}|^\frac{1}{2}}{\beta d_{min}(1-\gamma)} + \frac {\alpha}{\beta} \ln(|\mathcal{A}|).
\end{align*}
  
Then, applying the inequality recursively and letting $Q_0^U=Q_0^L$ produce
\begin{align*}
    & \mathbb{E}\left[\left\|Q_k^U-Q_k^L\right\|_{\infty}\right] \\
    \le& 2\alpha\gamma d_{max}|\mathcal{S} \times \mathcal{A}|^\frac{1}{2} \left\|Q_0^L-Q^*\right\|_2 k \rho^{k-1} \\
    &+ \frac{2\sqrt{6} \alpha^\frac{3}{2}\gamma d_{max} \left(\ln (|\mathcal{A}|)+\beta\right) |\mathcal{S} \times \mathcal{A}|^\frac{1}{2}}{\beta d_{min}^\frac{1}{2}(1-\gamma)^\frac{3}{2}} \sum_{i=0}^{k-1} \rho^{i} \\
    &+ \frac {2\alpha \gamma^2 d_{max}^2 \ln(|\mathcal{A}|) |\mathcal{S} \times \mathcal{A}|^\frac{1}{2}}{\beta d_{min}(1-\gamma)} \sum_{i=0}^{k-1} \rho^{i} + \frac{\alpha\ln(|\mathcal{A}|)}{\beta} \sum_{i=0}^{k-1} \rho^{i}.
\end{align*}

Noting the relation
\begin{equation} \label{Q_0^L-Q^* inequality 2}
\begin{aligned}
    \left\|Q_0^L-Q^*\right\|_2 \le& |\mathcal{S} \times \mathcal{A}|^\frac{1}{2} \left\|Q_0^L-Q^*\right\|_{\infty} \\
    \le& |\mathcal{S} \times \mathcal{A}|^\frac{1}{2} \left( \left\|Q_0^L\right\|_{\infty} + \left\|Q^*\right\|_{\infty} \right) \\
    \le& |\mathcal{S} \times \mathcal{A}|^\frac{1}{2} \left(1+\frac{1}{1-\gamma} \right) \\
    \le& |\mathcal{S} \times \mathcal{A}|^\frac{1}{2} \frac{2}{1-\gamma}
\end{aligned}
\end{equation}
and $\sum_{i=0}^{k-1} \rho^{i} \le \sum\limits_{i = 0}^\infty  {{\rho ^i}} \le \frac{1}{1 - \rho}$ with~\cref{as: 1,lem3.6,def3}, the last inequality turns into
\begin{align*}
    & \mathbb{E}\left[\left\|Q_k^U-Q_k^L\right\|_{\infty}\right] \\
    \le& \frac{4\alpha\gamma d_{max} |\mathcal{S} \times \mathcal{A}|}{1-\gamma} k \rho^{k-1} \\
    &+ \frac{2\sqrt{6} \alpha^\frac{1}{2}\gamma d_{max} \left(\ln (|\mathcal{A}|)+\beta\right) |\mathcal{S} \times \mathcal{A}|^\frac{1}{2}}{\beta d_{min}^\frac{3}{2}(1-\gamma)^\frac{5}{2}} \\
    &+ \frac {2\gamma^2 d_{max}^2 \ln(|\mathcal{A}|) |\mathcal{S} \times \mathcal{A}|^\frac{1}{2}}{\beta d_{min}^2(1-\gamma)^2} + \frac{\ln(|\mathcal{A}|)}{{\beta {d_{\min }}(1 - \gamma )}}.
\end{align*}

Combining the above inequality with the relation
\begin{align*}
    & \mathbb{E}\left[\left\|Q_k^{Boltz}-Q^*\right\|_{\infty}\right] \\
    =& \mathbb{E}\left[\left\|Q_k^{Boltz}-Q_k^L+Q_k^L-Q^*\right\|_{\infty}\right] \\
    \le& \mathbb{E}\left[\left\|Q_k^{Boltz}-Q_k^L\right\|_{\infty}\right]+ \mathbb{E}\left[\left\|Q_k^L-Q^*\right\|_{\infty}\right] \\
    \le& \mathbb{E}\left[\left\|Q_k^U-Q_k^L\right\|_{\infty}\right]+ \mathbb{E}\left[\left\|Q_k^L-Q^*\right\|_{\infty}\right]
\end{align*}
and~\cref{thm3.3} yield the desired conclusion as follows:
\begin{align*}
    & \mathbb{E}\left[\left\|Q_k^{boltz}-Q^*\right\|_{\infty}\right] \\
    \le& \mathbb{E}\left[\left\|Q_k^U-Q_k^L\right\|_{\infty}\right]+ \mathbb{E}\left[\left\|Q_k^L-Q^*\right\|_{\infty}\right] \\
    \le& \frac{4\alpha\gamma d_{max} |\mathcal{S} \times \mathcal{A}|}{1-\gamma} k \rho^{k-1} \\
    &+ \frac{2\sqrt{6} \alpha^\frac{1}{2}\gamma d_{max} \left(\ln (|\mathcal{A}|)+\beta\right) |\mathcal{S} \times \mathcal{A}|^\frac{1}{2}}{\beta d_{min}^\frac{3}{2}(1-\gamma)^\frac{5}{2}} \\
    &+ \frac {2\gamma^2 d_{max}^2 \ln(|\mathcal{A}|) |\mathcal{S} \times \mathcal{A}|^\frac{1}{2}}{\beta d_{min}^2(1-\gamma)^2} + \frac{\ln(|\mathcal{A}|)}{{\beta {d_{\min }}(1 - \gamma )}} \\
    &+ |\mathcal{S} \times \mathcal{A}|^\frac{1}{2} \|Q_0^L-Q^*\|_2 \rho^{k} \\
    &+ \frac{\sqrt{6} \alpha^\frac{1}{2} \left(\ln (|\mathcal{A}|)+\beta\right) |\mathcal{S} \times \mathcal{A}|^\frac{1}{2}}{\beta d_{min}^\frac{1}{2}(1-\gamma)^\frac{3}{2}} \\
    &+ \frac {\gamma d_{max} \ln(|\mathcal{A}|)|\mathcal{S} \times \mathcal{A}|^\frac{1}{2}}{\beta d_{min}(1-\gamma)} \\
    \le& \frac{4\alpha\gamma d_{max} |\mathcal{S} \times \mathcal{A}|}{1-\gamma} k \rho^{k-1} \\
    &+ \frac{2\sqrt{6} \alpha^\frac{1}{2}\gamma d_{max} \left(\ln (|\mathcal{A}|)+\beta\right) |\mathcal{S} \times \mathcal{A}|^\frac{1}{2}}{\beta d_{min}^\frac{3}{2}(1-\gamma)^\frac{5}{2}} \\
    &+ \frac {2\gamma^2 d_{max}^2 \ln(|\mathcal{A}|) |\mathcal{S} \times \mathcal{A}|^\frac{1}{2}}{\beta d_{min}^2(1-\gamma)^2} + \frac{\ln(|\mathcal{A}|)}{{\beta {d_{\min }}(1 - \gamma )}} \\
    &+ \frac{2 |\mathcal{S} \times \mathcal{A}|}{1-\gamma} \rho^k + \frac{\sqrt{6} \alpha^\frac{1}{2} \left(\ln (|\mathcal{A}|)+\beta\right) |\mathcal{S} \times \mathcal{A}|^\frac{1}{2}}{\beta d_{min}^\frac{1}{2}(1-\gamma)^\frac{3}{2}} \\
    &+ \frac {\gamma d_{max} \ln(|\mathcal{A}|)|\mathcal{S} \times \mathcal{A}|^\frac{1}{2}}{\beta d_{min}(1-\gamma)} \\
    \le& \frac{4\alpha\gamma d_{max} |\mathcal{S} \times \mathcal{A}|}{1-\gamma} k \rho^{k-1} \\
    &+ \frac{3\sqrt{6} \alpha^\frac{1}{2} d_{max} \left(\ln (|\mathcal{A}|)+\beta\right) |\mathcal{S} \times \mathcal{A}|^\frac{1}{2}}{\beta d_{min}^\frac{3}{2}(1-\gamma)^\frac{5}{2}} \\
    &+ \frac {4 d_{max} \ln(|\mathcal{A}|) |\mathcal{S} \times \mathcal{A}|^\frac{1}{2}}{\beta d_{min}^2(1-\gamma)^2} + \frac{2 |\mathcal{S} \times \mathcal{A}|}{1-\gamma} \rho^k,
\end{align*}
where~\eqref{Q_0^L-Q^* inequality 2} is used in the third inequality, and $\gamma \in [0,1)$,~\cref{def1} is utilized in the last inequality.
\end{proof}

\end{document}